\documentclass[10pt,twocolumn,letterpaper]{article}

\usepackage{cvpr}
\usepackage{multirow}
\usepackage{tabularx}
\usepackage{array}
\definecolor{cvprblue}{rgb}{0.21,0.49,0.74}
\usepackage[pagebackref,breaklinks,colorlinks,allcolors=cvprblue]{hyperref}

\def\paperID{}
\def\confName{CVPR}
\def\confYear{2026}

\title{Fine-Grained Multi-Image Object Hallucination Benchmark}

\author{
Joonki Min$^{1{}*}$,
Chaeyun Kim$^{1,2{}*}$,
Hyungwook Choi$^{1}$,
Yejin Kim$^{1}$,
Kihyun Kim$^{1}$ \\
Yohan Jo$^{1{}\dagger}$,
Joonseok Lee$^{1{}\dagger}$\\
$^{1}$Seoul National University \qquad $^{2}$AIM Intelligence\\
{\tt\small \{minc69,golddohyun,chooi221,a2000yejin,ki5477,yohan.jo,joonseok\}@snu.ac.kr}
}

\begin{document}
\maketitle
\renewcommand{\thefootnote}{\fnsymbol{footnote}}
\footnotetext[1]{Equal contribution.}
\footnotetext[2]{Corresponding author.}
\renewcommand{\thefootnote}{\arabic{footnote}}

\begin{abstract}
Multimodal Large Language Models (MLLMs) are increasingly deployed in multi-image scenarios requiring complex reasoning across visual contexts.
However, current MLLMs remain fundamentally limited by object hallucination—generating plausible yet factually inconsistent descriptions about objects.
Existing benchmarks, designed primarily for single-image settings or providing only high-level multi-image assessments, cannot systematically diagnose how visual complexity and reasoning demands trigger hallucination.
To address this gap, we introduce MIOH, a fine-grained
multi-image object hallucination benchmark that systematically evaluates object hallucination across four foundational tasks (existence, counting, attribute, position) through three multi-image reasoning patterns (comprehensive, comparative, selective) under three controlled adversarial pressures (visual context scale, perceptual difficulty, contextual bias).
Through evaluation of 29 models, we reveal that even state-of-the-art systems like GPT-5 and Gemini-2.5-Pro exhibit distinct failure patterns across different reasoning patterns and tasks.
Our evaluation reveals that hallucination stems not merely from perceptual failures but from integration-stage limitations when maintaining object representations across multiple images.
MIOH provides a controlled framework for analyzing multi-image object hallucination and serves as a critical evaluation tool for developing more reliable multimodal AI systems.
\end{abstract}

\section{Introduction}
\label{sec:intro}

With recent advances, Multimodal Large Language Models (MLLMs) are capable of reasoning over multiple images simultaneously.
This capability requires models not only to recognize content in individual images but also to integrate and synthesize information across diverse visual inputs.
Despite these advancements, however, current MLLMs remain fundamentally limited by \textit{object hallucination}, where models generate plausible yet factually inconsistent descriptions about objects in the queried images.

Multi-image scenarios amplify object hallucination along two critical dimensions. First, \textbf{\emph{perceptual failure}} : even within individual images, models struggle with visually challenging objects (small size, occlusion, contextual ambiguity) or contextually misleading scenes where co-occurring objects create false expectations. Second, \textbf{\emph{information integration}} : multi-image contexts demand integrating and synthesizing information across images, where models must aggregate information across all images \textbf{(comprehensive reasoning)}, identify differences between images \textbf{(comparative reasoning)}, or retrieve specific images matching given criteria \textbf{(selective reasoning)}. Failures in either dimension create compounding pathways for hallucination. Systematic evaluation requires isolating both: which visual conditions trigger perceptual errors, and which integration demands are most vulnerable to failure.

Despite the importance of understanding object hallucination in multi-image contexts, existing evaluation frameworks do not adequately address these two dimensions. Most object hallucination benchmarks \cite{li2023evaluating, rohrbach2018object, throne, mmhal-bench, amber} are designed for single-image scenarios with binary questions, focusing narrowly on existence and counting.
This design cannot reveal how visual difficulty factors (perceptual challenges, contextual biases) systematically induce hallucination, nor can it assess multi-image reasoning patterns or compositional capabilities (attributes, spatial relations) that require fine-grained object understanding.
While general multi-image benchmarks \citep{wang2024muirbench, meng2024mmiu, liu2024mmdu, jiang2024mantis, blink, demon} evaluate overall reasoning, they lack the controlled manipulation of visual factors needed to diagnose object hallucination vulnerabilities.
Recent work such as MIHBench \citep{li2025mihbench} explores multi-image hallucination but examines only image quantity as an adversarial factor in ablation studies, without systematically varying visual difficulty or distinguishing between reasoning patterns.
Consequently, existing benchmarks cannot pinpoint which specific visual conditions or reasoning demands trigger hallucination, nor how these factors interact.

To narrow these gaps, we introduce the Fine-Grained Multi-Image Object Hallucination (MIOH) benchmark, designed to systematically evaluate both dimensions of \textbf{multi-image object hallucination}.
MIOH systematically integrates four object-centric tasks (existence, counting, attribute, position)
with three reasoning patterns (comprehensive, comparative, selective), enabling compositional evaluation from basic detection to fine-grained property binding.
Furthermore, MIOH introduces three controlled adversarial pressures—visual context scale (number of images), perceptual difficulty (small/occluded objects), and contextual bias (misleading co-occurrence priors)—that systematically vary visual complexity while measuring performance across reasoning patterns.
This design enables diagnostic analysis by separately measuring performance under
different visual conditions (perceptual difficulty, contextual bias, image scale)
and across different reasoning patterns (comprehensive, comparative, selective).

Our contributions are summarized as follows:

\begin{itemize}[leftmargin=5mm, topsep=0pt]
    \setlength{\itemsep}{0pt}
    \setlength{\parskip}{0pt}

    \item We introduce MIOH, the first benchmark to systematically assess
    \textbf{object hallucination in multi-image contexts} with fine-grained
    diagnostic capabilities across visual complexity and reasoning demands.

    \item We define \textbf{three multi-image reasoning patterns} (comprehensive,
    comparative, selective) and instantiate them across four foundational
    object-centric tasks (existence, counting, attribute, position), enabling
    diagnosis of which reasoning capability fails under hallucination pressures.

    \item We design \textbf{three controllable adversarial pressures} (visual
    context scale, perceptual difficulty, contextual bias) that systematically
    exacerbate hallucination, allowing fine-grained analysis of when and why
    MLLMs fail in multi-image scenarios.
\end{itemize}

\section{Related Work}
\label{sec:related}

\textbf{Multimodal Large Language Models (MLLMs).}
Following the success of LLMs, MLLMs have rapidly evolved through visual instruction tuning, utilized by LLaVA \citep{liu2023visual} and extended by InstructBLIP \citep{dai2023instructblip} and MiniGPT-4 \citep{zhu2023minigpt}.
Early MLLMs face challenges in cross-image reasoning due to limitations in visual token processing and inter-image semantic modeling \citep{li2023blip, dai2023instructblip,cha2024honeybee,alayrac2022flamingo,lee2026llm}.
Recent work has enabled multi-image understanding \citep{jiang2024mantis, li2024llava,laurenccon2024building,yao2024minicpm,lu2025ovis2};
Mantis \citep{jiang2024mantis}, LLaVA-NeXT-Interleave \citep{li2024llava}, and Idefics3 \citep{laurenccon2024building} leverage large-scale image-text data at training, and Qwen2.5-VL \citep{bai2025qwen2}, InternVL3.5 \citep{wang2025internvl3}, and Gemini-2.5-Pro \citep{comanici2025gemini} demonstrate powerful cross-image reasoning capabilities.

\vspace{0.1cm} \noindent
\textbf{Object Hallucination in MLLMs.}
Object hallucination, defined as MLLMs generating plausible but inaccurate object descriptions inconsistent with visual inputs, remains a critical challenge \citep{rohrbach2018object, dai2022plausible}.
Systematic analysis has pinpointed causes across the MLLM pipeline:
data-related issues such as statistical biases in training data \citep{li2023evaluating},
limitations of vision encoder in fine-grained semantics \citep{tong2024eyes}, insufficient modality alignment \citep{liu2024improved}, and inherited LLM biases such as weak context attention \citep{wang2024cogvlm}.
Recent studies reveal additional visual vulnerabilities, \emph{e.g.},
perception of small or occluded objects \citep{zhang2024exploring}, contextual bias due to object co-occurrence patterns \citep{li2023evaluating} and semantic similarities \citep{li2024naturalbench}.
MLLMs are also linguistically susceptible to sycophantic alignment with user beliefs and context hijacking from misleading narratives \citep{zhao2024towards}.
Mitigation strategies, \emph{e.g.}, data augmentation \citep{sarkar2024mitigating}, preference optimization \citep{zhang2024automated}, and inference-time interventions \citep{zhaomitigating, he2025evaluating}, have primarily targeted single image scenarios.
Multi-image contexts amplify hallucination challenges, requiring models not only to recognize objects accurately, but to track them and maintain contextual consistency across distinct images \citep{wu2024visual}.
This increased complexity requires a new benchmark tailored to assess object hallucination on multiple images.

\vspace{0.1cm} \noindent
\textbf{Benchmarks for MLLMs.}
Early MLLM benchmarks focus on single-image scenarios across various tasks including visual question answering, reasoning, and compositional understanding \citep{visualvqa, fu2023mme, seed-bench, vsr, mmbench}.
Recently,
several benchmarks \citep{wang2024muirbench, meng2024mmiu, liu2024mmdu, jiang2024mantis, blink, demon}
assess general reasoning capabilities across multiple images, though none of them specifically target object hallucination.

\vspace{0.1cm} \noindent
In parallel, object hallucination has been addressed through specialized benchmarks, including discriminative approaches using binary questions \citep{li2023evaluating, hallusionbench} and generative approaches directly assessing free-form descriptions \citep{rohrbach2018object, throne, mmhal-bench, amber}.
They typically focus on existence and simple counting tasks, with limited coverage of attributes or spatial relations.
Also, they predominantly employ simple binary questions or captioning tasks, confined to single-image settings.
Recently, MIHBench\citep{li2025mihbench} has also explored hallucination in multi-image settings. However, it focuses mainly on binary existence questions and varies only the number of images, without modeling broader visual difficulty factors or diverse object-centric tasks.
Our benchmark is complementary to this direction, offering a more fine-grained diagnostic space across multiple tasks, reasoning patterns, and adversarial conditions.

\begin{figure*}[t]
  \centering
  \includegraphics[width=\textwidth]{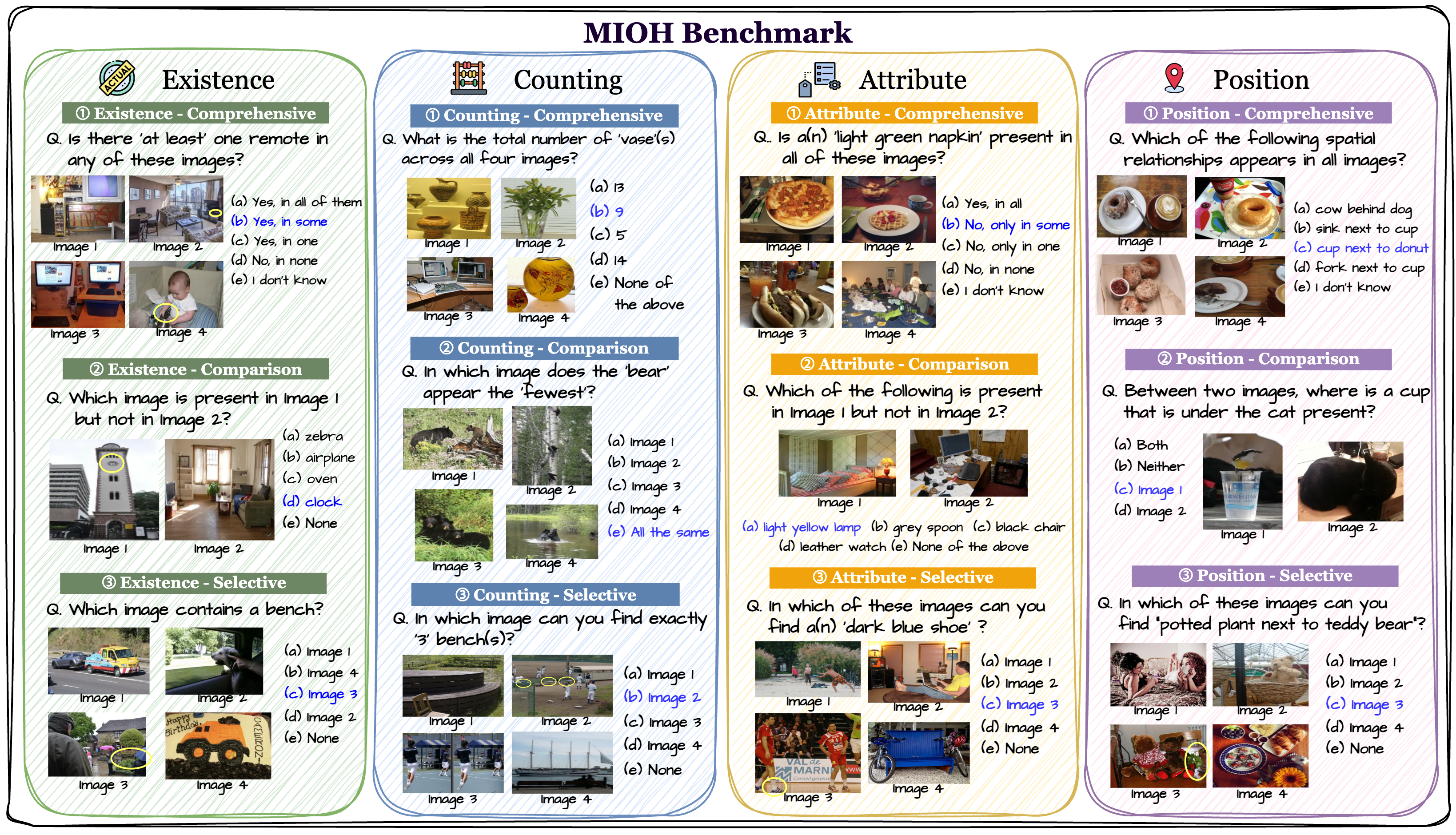}
  \caption{\textbf{Overview of our MIOH Benchmark.} MIOH evaluates object hallucination in multi-image contexts across four core tasks: existence, counting, attribute, and position. Each task includes three question types (comprehensive, comparative, and selective) designed to probe different aspects of multi-image reasoning capabilities.}
  \label{fig:benchmark_fig}
  \vspace{-0.3cm}
\end{figure*}

\section{Overall Design of our MIOH Benchmark}
\label{sec:method}

We introduce the Multi-Image Object Hallucination (MIOH) benchmark, designed to
systematically evaluate object hallucination in MLLMs under multi-image contexts.
MIOH is structured around two complementary dimensions: (1) \textbf{multi-image
reasoning patterns} that probe different integration capabilities
(\cref{sec:reasoning_patterns}), and (2) \textbf{adversarial pressures} that
systematically challenge perceptual and contextual robustness
(\cref{sec:adversarial_pressure}). By crossing four fundamental object-centric
tasks with three reasoning patterns under varying adversarial conditions, MIOH
enables fine-grained diagnosis of when and why MLLMs hallucinate.

\subsection{Multi-Image Reasoning Patterns}
\label{sec:reasoning_patterns}

\textbf{Motivation.} Multi-image understanding requires fundamentally different reasoning capabilities compared to single-image settings.
While single-image benchmarks test whether models can recognize objects in isolation, multi-image scenarios demand that models \emph{synthesize, compare, and retrieve} information across multiple visual contexts.
We identify three core reasoning patterns that capture these distinct capabilities, each placing unique demands on the model's
information integration mechanisms.

\vspace{0.1cm} \noindent
\textbf{Comprehensive Reasoning.} requires models to aggregate information across all images to form a holistic judgment.
Questions in this category ask about collective properties spanning the entire image set, such as ``Does any image contain a zebra?'' or ``What is the total count of cars across all images?''
This pattern tests the model's ability to maintain a unified representation of information distributed across multiple scenes—a capability essential for summarization and global understanding tasks.
Failures in comprehensive reasoning indicate difficulties in information aggregation or token limitations.

\vspace{0.1cm} \noindent
\textbf{Comparative Reasoning.} requires models to identify differences between specific images.
Questions such as ``Which image contains more cars?'' demand precise cross-image attention
and the ability to maintain separate representations for different scenes while comparing them.
This is critical for change detection, progress monitoring, and contrastive analysis.
Failures suggest weaknesses in maintaining distinct object representations across contexts.

\vspace{0.1cm} \noindent
\textbf{Selective Reasoning.} requires models to retrieve a specific image that matches given criteria from a set of candidates. Questions like ``In which image are exactly three zebras present?'' test the ability to perform targeted retrieval while filtering out irrelevant information.
This is essential for search and localization tasks in multi-image contexts.
Failures indicate poor image-level indexing or inability to isolate relevant visual evidence from distractors.

\vspace{0.1cm} \noindent
\textbf{Diagnostic Value.} These three patterns are not merely different question formats—they probe distinct failure modes.
Uniform failure across all patterns suggests a fundamental bottleneck in object recognition (perceptual failure).
Conversely, failure in \textit{selective} reasoning while succeeding in \textit{comprehensive} reasoning points to a specific integration failure in retrieval and localization.
This systematic evaluation allows us to diagnose \emph{which} integration capability breaks down under specific adversarial conditions.

\begin{figure*}[t]
  \centering
  \includegraphics[width=\textwidth]{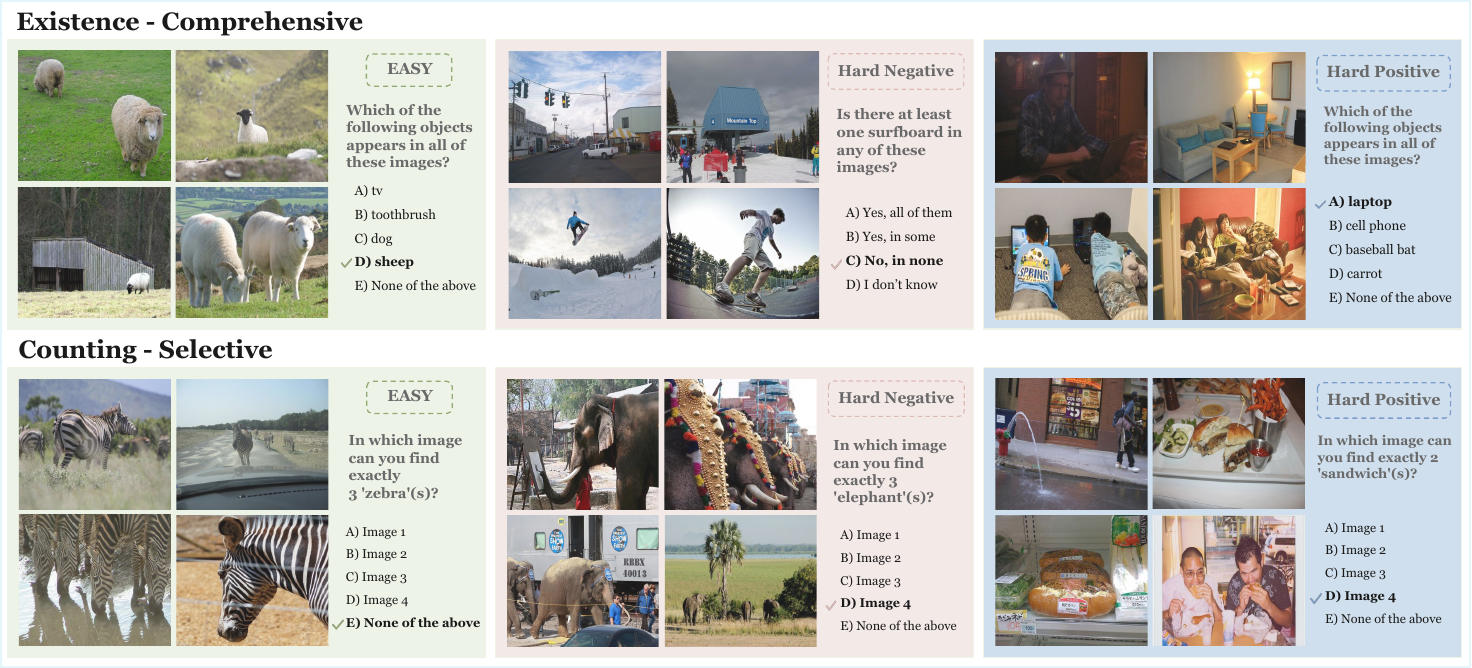}
  \caption{\textbf{Representative examples from MIOH.} (Top) Existence-Comprehensive questions across Easy, Hard Negative, and Hard Positive scenarios. (Bottom) Counting-Selective questions with varying difficulty levels. Additional examples are provided in \cref{appendix:examples}.}
  \label{fig:benchmark_samples}
  \vspace{-0.5cm}
\end{figure*}

\subsection{Object-Centric Tasks}
\label{sec:core_tasks}

We apply these three reasoning patterns to four fundamental object-centric tasks
that represent core visual understanding capabilities: \textbf{Existence} (verifying object presence/absence), \textbf{Counting} (enumerating objects), \textbf{Attribute} (binding visual properties to objects, e.g., ``red car''), and \textbf{Position} (binding spatial relationships, e.g., ``dog next to a cat'').
While Existence and Counting have been the primary focus of object hallucination benchmarks \citep{rohrbach2018object, li2023evaluating, lovenia2023negative, throne, wang2024mitigating, chen2024multi}, all four tasks form the foundation for assessing object-centric capabilities of MLLMs \citep{fu2023mme, mmbench, jing2024faithscore, mmhal-bench, qiu2024longhalqa,
wang2024muirbench, villa2025behind}.

\vspace{0.1cm} \noindent
\textbf{From Tasks to Question Types.} By crossing four tasks with three reasoning patterns, we create a comprehensive evaluation space. However, not all combinations are equally meaningful or distinct.
For example, in the Counting task, comprehensive reasoning (``total count across all images'') naturally leads to multiple variants depending on whether we ask for exact counts, comparisons of counts, or presence of specific quantities.
Through careful design, we developed \textbf{26 distinct question types} that systematically cover the task and reasoning space while ensuring each type probes a unique aspect of multi-image understanding.
~\cref{tab:question_types} presents the complete taxonomy with representative templates for each type. Complete examples are in ~\cref{fig:benchmark_fig} and ~\cref{appendix:examples}.

\begin{table}
\centering
\scriptsize
\caption{\textbf{MIOH Question Type Taxonomy.} We design 26 question
types by crossing 4 object-centric tasks with 3 reasoning patterns. Each type is illustrated with a template showing its structure.}
\vspace{-0.2cm}
\label{tab:question_types}

\renewcommand{\arraystretch}{0.90}
\begin{tabularx}{\linewidth}{@{}l@{\hspace{1mm}}X@{}}
\toprule
\textbf{Type} & \textbf{Template} \\
\midrule

\multicolumn{2}{@{}c}{\textbf{Existence}} \\
\midrule

Comprehensive & Is there at least one \{object\} in any of these images? \\
Comprehensive & Is a \{object\} present in all of these images? \\
Comprehensive  & Which object appears in at least one image? \\
Comprehensive  & Which object appears in all images? \\
Selective     & In which image does a \{object\} appear? \\
Comparative   & Which of the two images contains a \{object\}? \\
Comparative   & Which object is in Image~1 but not Image~2? \\

\midrule

\multicolumn{2}{@{}c}{\textbf{Attribute}} \\
\cmidrule{1-2}

Comprehensive & Is a ``\{attribute\} \{object\}'' present in any image? \\
Comprehensive & Is a ``\{attribute\} \{object\}'' present in all images? \\
Comprehensive    & Which attribute--object pair appears in one image? \\
Comprehensive   & Which attribute--object pair appears in all images? \\
Selective     & In which image is ``\{attribute\} \{object\}'' found? \\
Comparative   & Which of the two images ``\{attribute\} \{object\}'' present? \\
Comparative   & Which attribute--object pair is in Image~1 but not Image~2? \\

\midrule

\multicolumn{2}{@{}c}{\textbf{Counting}} \\
\cmidrule{1-2}

Comprehensive & What is the total number of ``\{object\}'' across images? \\
Comprehensive & Which object appears \{count\} times across images? \\
Comprehensive   & In how many images is a ``\{object\}'' present? \\
Selective     & In which image are \{count\} ``\{object\}'' found? \\
Comparative    & In which image does the ``\{object\}'' appear most/least? \\

\midrule

\multicolumn{2}{@{}c}{\textbf{Position}} \\
\cmidrule{1-2}

Comprehensive & Is a \{subject\} \{relation\} a \{anchor\} present in any image? \\
Comprehensive & Is a \{subject\} \{relation\} a \{anchor\} present in all images? \\
Comprehensive & Which object with spatial relationship appears in one image? \\
Comprehensive & Which object with spatial relationship appears in all images? \\
Selective     & In which image is a \{subject\} \{relation\} a \{anchor\}? \\
Selective   & Where is ``\{subject\} \{relation\} \{anchor\}'' present? \\
Comparative   & Which object with spatial relationship is in Image~1 but not Image~2? \\

\bottomrule
\end{tabularx}
\vspace{-0.3cm}
\end{table}

\renewcommand{\arraystretch}{0.70}
\begin{table*}[h!]
    \centering
    \scriptsize
    \resizebox{\textwidth}{!}{
        \begin{tabular}{l ccccc c ccccc c}
            \toprule[1.2pt]
            \multirow{2}{*}{\textbf{Model}} & \multicolumn{5}{c}{\textbf{Existence}} & \multicolumn{5}{c}{\textbf{Counting}} \\
            \cmidrule(lr){2-6} \cmidrule(lr){7-11}
            & Easy & HN & HP & NI & \textbf{Avg} & Easy & HN & HP & NI & \textbf{Avg} \\
            \midrule
            \textbf{Overall} & 62.4 & 51.9 & 51.5 & 30.0 & \textbf{49.4} & 24.3 & 29.5 & 27.0 & 20.5 & \textbf{25.4} \\
            \midrule
            \textbf{GPT-5} & 91.4 & 88.6 & 78.4 & 55.3 & \textbf{78.4} & 55.0 & 54.3 & 51.5 & 35.7 & \textbf{49.1} \\
            \textbf{Gemini-2.5 Pro} & 83.0 & 81.5 & 80.7 & 56.4 & \textbf{75.4} & 64.4 & 66.4 & 59.4 & 40.0 & \textbf{57.5} \\
            \midrule
            \textbf{Qwen2-VL-2B} & 58.3 & 53.5 & 45.9 & 32.7 & \textbf{47.6} & 21.7 & 23.3 & 18.6 & 21.4 & \textbf{21.2} \\
            \textbf{Qwen2.5-VL-3B} & 80.7 & 62.5 & 55.5 & 44.7 & \textbf{60.8} & 23.3 & 26.7 & 19.6 & 17.1 & \textbf{21.7} \\
            \textbf{Qwen2-VL-7B} & 87.3 & 75.5 & 76.4 & 30.7 & \textbf{67.5} & 28.9 & 36.2 & 21.6 & 17.1 & \textbf{26.0} \\
            \textbf{Qwen2.5-VL-7B} & 84.0 & 76.0 & 73.6 & 28.0 & \textbf{65.4} & 33.3 & 39.7 & 17.5 & 21.4 & \textbf{28.0} \\
            \midrule
            \textbf{LLaVA-v1.6 (Mistral-7B)} & 33.0 & 47.0 & 36.8 & - & \textbf{38.9} & 20.6 & 27.6 & 18.6 & - & \textbf{22.2} \\
            \textbf{LLaVA-Interleave (Qwen-0.5B)} & 36.3 & 31.5 & 28.6 & 18.7 & \textbf{28.8} & 22.8 & 28.4 & 24.7 & 14.3 & \textbf{22.6} \\
            \textbf{LLaVA-Interleave (Qwen-7B)} & 46.7 & 44.0 & 65.5 & 30.7 & \textbf{46.7} & 20.6 & 25.0 & 28.9 & 24.3 & \textbf{24.7} \\
            \textbf{LLaVA-Interleave (Qwen-7B-DPO)} & 67.7 & 44.0 & 55.9 & 31.3 & \textbf{49.7} & 21.1 & 25.9 & 30.9 & 27.1 & \textbf{26.3} \\
            \textbf{LLaVA-OneVision (Qwen2-0.5B-SI)} & 27.7 & 11.5 & 37.3 & 40.0 & \textbf{29.1} & 13.9 & 21.6 & 18.6 & 12.9 & \textbf{16.7} \\
            \textbf{LLaVA-OneVision (Qwen2-0.5B-OV)} & 57.3 & 23.5 & 33.6 & 22.0 & \textbf{34.1} & 16.7 & 26.7 & 16.5 & 18.6 & \textbf{19.6} \\
            \textbf{LLaVA-OneVision (Qwen2-7B-SI)} & 68.3 & 46.5 & 40.9 & 36.0 & \textbf{47.9} & 17.2 & 24.1 & 25.8 & 25.7 & \textbf{23.2} \\
            \textbf{LLaVA-OneVision (Qwen2-7B-OV)} & 83.3 & 78.0 & 57.7 & 24.7 & \textbf{60.9} & 18.3 & 28.4 & 29.9 & 18.6 & \textbf{23.8} \\
            \textbf{LLaVA-OneVision (Qwen2-7B-OV-Chat)} & 82.7 & 77.0 & 43.6 & 26.7 & \textbf{57.5} & 20.0 & 30.2 & 27.8 & 21.4 & \textbf{24.9} \\
            \midrule
            \textbf{InternVL3.5-1B} & 44.0 & 41.0 & 30.5 & 42.0 & \textbf{39.4} & 21.7 & 21.6 & 19.6 & 21.4 & \textbf{21.1} \\
            \textbf{InternVL3.5-2B} & 51.7 & 37.5 & 36.8 & 26.0 & \textbf{38.0} & 15.0 & 28.4 & 16.5 & 17.1 & \textbf{19.3} \\
            \textbf{InternVL3.5-4B} & 59.0 & 71.0 & 60.0 & 22.7 & \textbf{53.2} & 18.9 & 27.6 & 34.0 & 17.1 & \textbf{24.4} \\
            \textbf{InternVL3.5-8B Pretrained} & 72.0 & 62.0 & 59.5 & 30.0 & \textbf{55.9} & 22.8 & 26.7 & 27.8 & 14.3 & \textbf{22.9} \\
            \textbf{InternVL3.5-8B Instruct} & 76.0 & 59.5 & 75.5 & 30.7 & \textbf{60.4} & 25.6 & 29.3 & 29.9 & 8.6 & \textbf{23.3} \\
            \textbf{InternVL3.5-8B MPO} & 76.7 & 60.0 & 76.4 & 18.0 & \textbf{57.8} & 26.1 & 29.3 & 28.9 & 10.0 & \textbf{23.6} \\
            \textbf{InternVL3.5-8B} & 34.0 & 55.5 & 34.1 & 20.0 & \textbf{35.9} & 30.0 & 37.1 & 32.0 & 11.4 & \textbf{27.6} \\
            \midrule
            \textbf{Mantis-8B (CLIP-Llama3)} & 45.7 & 57.0 & 36.8 & 18.7 & \textbf{39.5} & 22.8 & 25.9 & 21.6 & 30.0 & \textbf{25.1} \\
            \textbf{Mantis-8B (SIGLIP-Llama3)} & 47.0 & 45.0 & 42.7 & 24.7 & \textbf{39.8} & 20.6 & 25.0 & 33.0 & 22.9 & \textbf{25.4} \\
            \textbf{MiniCPM-Llama3-V-2.5} & 35.0 & 16.5 & 25.5 & 4.7 & \textbf{20.4} & 4.4 & 5.2 & 8.2 & 5.7 & \textbf{5.9} \\
            \textbf{MiniCPM-V-2.6} & 86.3 & 75.5 & 73.6 & 38.7 & \textbf{68.5} & 21.1 & 35.3 & 38.1 & 18.6 & \textbf{28.3} \\
            \textbf{Ovis2.5-2B} & 72.0 & 23.5 & 47.3 & 34.0 & \textbf{44.2} & 23.9 & 25.9 & 18.6 & 25.7 & \textbf{23.5} \\
            \textbf{Ovis2.5-9B} & 78.0 & 23.0 & 51.8 & 28.7 & \textbf{45.4} & 28.9 & 27.6 & 48.5 & 25.7 & \textbf{32.7} \\
            \textbf{Phi-4-multimodal} & 45.3 & 38.0 & 33.6 & 23.3 & \textbf{35.1} & 25.0 & 25.0 & 17.5 & 28.6 & \textbf{24.0} \\
            \bottomrule[1.2pt]
        \end{tabular}
    }

    \resizebox{\textwidth}{!}{
        \begin{tabular}{l ccccc c ccccc c c}
            \toprule[1.2pt]
            \multirow{2}{*}{\textbf{Model}} & \multicolumn{5}{c}{\textbf{Attribute}} & \multicolumn{5}{c}{\textbf{Position}} & \multirow{2}{*}{\textbf{Overall}} \\
            \cmidrule(lr){2-6} \cmidrule(lr){7-11}
            & Easy & HN & HP & NI & \textbf{Avg} & Easy & HN & HP & NI & \textbf{Avg} & \textbf{Avg} \\
            \midrule
            \textbf{Overall} & 37.5 & 32.9 & 31.3 & 26.9 & \textbf{32.4} & 46.9 & 35.2 & 45.2 & 22.2 & \textbf{37.3} & \textbf{36.1} \\
            \midrule
            \textbf{GPT-5} & 65.8 & 62.4 & 57.0 & 45.8 & \textbf{57.8} & 88.5 & 70.3 & 75.5 & 34.1 & \textbf{67.1} & \textbf{63.1} \\
            \textbf{Gemini-2.5 Pro} & 62.9 & 57.1 & 64.3 & 47.5 & \textbf{57.9} & 80.8 & 66.4 & 81.4 & 37.6 & \textbf{66.6} & \textbf{64.4} \\
            \midrule
            \textbf{Qwen2-VL-2B} & 42.3 & 34.2 & 50.0 & 25.0 & \textbf{37.9} & 46.2 & 38.4 & 74.4 & 22.0 & \textbf{45.3} & \textbf{38.0} \\
            \textbf{Qwen2.5-VL-3B} & 47.3 & 38.8 & 47.1 & 26.7 & \textbf{40.0} & 73.8 & 50.4 & 72.1 & 22.0 & \textbf{54.6} & \textbf{44.3} \\
            \textbf{Qwen2-VL-7B} & 58.8 & 36.2 & 44.3 & 28.3 & \textbf{41.9} & 86.2 & 56.8 & 79.5 & 21.7 & \textbf{61.1} & \textbf{49.1} \\
            \textbf{Qwen2.5-VL-7B} & 52.7 & 42.1 & 38.1 & 26.7 & \textbf{39.9} & 76.2 & 52.4 & 60.5 & 21.4 & \textbf{52.6} & \textbf{46.5} \\
            \midrule
            \textbf{LLaVA-v1.6 (Mistral-7B)} & 30.4 & 21.7 & 24.3 & - & \textbf{25.4} & 29.2 & 25.2 & 31.2 & - & \textbf{28.5} & \textbf{28.8} \\
            \textbf{LLaVA-Interleave (Qwen-0.5B)} & 41.5 & 31.7 & 31.4 & 17.5 & \textbf{30.5} & 15.0 & 17.6 & 40.9 & 34.5 & \textbf{27.0} & \textbf{27.2} \\
            \textbf{LLaVA-Interleave (Qwen-7B)} & 27.3 & 27.5 & 29.0 & 15.8 & \textbf{24.9} & 31.2 & 24.8 & 32.1 & 21.4 & \textbf{27.4} & \textbf{30.9} \\
            \textbf{LLaVA-Interleave (Qwen-7B-DPO)} & 28.5 & 28.8 & 28.6 & 13.3 & \textbf{24.8} & 37.1 & 26.4 & 30.2 & 18.5 & \textbf{28.1} & \textbf{32.2} \\
            \textbf{LLaVA-OneVision (Qwen2-0.5B-SI)} & 22.3 & 17.9 & 27.1 & 12.5 & \textbf{20.0} & 18.3 & 15.6 & 34.9 & 16.8 & \textbf{21.4} & \textbf{21.8} \\
            \textbf{LLaVA-OneVision (Qwen2-0.5B-OV)} & 23.1 & 17.1 & 28.1 & 22.5 & \textbf{22.7} & 19.2 & 21.2 & 33.5 & 22.9 & \textbf{24.2} & \textbf{25.2} \\
            \textbf{LLaVA-OneVision (Qwen2-7B-SI)} & 24.2 & 21.2 & 24.3 & 28.3 & \textbf{24.5} & 37.5 & 30.4 & 42.8 & 27.1 & \textbf{34.4} & \textbf{32.5} \\
            \textbf{LLaVA-OneVision (Qwen2-7B-OV)} & 30.4 & 24.6 & 27.1 & 23.3 & \textbf{26.4} & 47.9 & 24.8 & 43.7 & 21.1 & \textbf{34.4} & \textbf{36.4} \\
            \textbf{LLaVA-OneVision (Qwen2-7B-OV-Chat)} & 30.8 & 25.0 & 26.7 & 22.5 & \textbf{26.2} & 48.3 & 25.6 & 43.3 & 21.0 & \textbf{34.5} & \textbf{35.8} \\
            \midrule
            \textbf{InternVL3.5-1B} & 28.8 & 24.2 & 27.1 & 17.5 & \textbf{24.4} & 57.1 & 37.2 & 25.1 & 32.7 & \textbf{38.0} & \textbf{30.7} \\
            \textbf{InternVL3.5-2B} & 40.4 & 37.9 & 15.2 & 35.0 & \textbf{32.1} & 60.0 & 43.6 & 13.5 & 18.4 & \textbf{33.9} & \textbf{30.8} \\
            \textbf{InternVL3.5-4B} & 42.3 & 39.2 & 34.8 & 35.8 & \textbf{38.0} & 31.7 & 18.8 & 57.2 & 19.2 & \textbf{31.7} & \textbf{36.8} \\
            \textbf{InternVL3.5-8B Pretrained} & 40.8 & 37.5 & 19.5 & 38.3 & \textbf{34.0} & 37.5 & 32.0 & 37.2 & 28.1 & \textbf{33.7} & \textbf{36.6} \\
            \textbf{InternVL3.5-8B Instruct} & 38.1 & 34.6 & 18.1 & 39.2 & \textbf{32.5} & 35.4 & 33.2 & 37.7 & 19.8 & \textbf{31.5} & \textbf{36.9} \\
            \textbf{InternVL3.5-8B MPO} & 37.7 & 42.9 & 14.8 & 39.2 & \textbf{33.6} & 36.2 & 34.0 & 20.0 & 18.5 & \textbf{27.2} & \textbf{35.5} \\
            \textbf{InternVL3.5-8B} & 34.6 & 27.5 & 28.6 & 37.5 & \textbf{32.0} & 55.8 & 36.4 & 35.8 & 35.0 & \textbf{40.8} & \textbf{34.1} \\
            \midrule
            \textbf{Mantis-8B (CLIP-Llama3)} & 24.6 & 24.6 & 24.8 & 20.0 & \textbf{23.5} & 25.8 & 25.2 & 31.6 & 20.6 & \textbf{25.8} & \textbf{28.5} \\
            \textbf{Mantis-8B (SIGLIP-Llama3)} & 26.2 & 29.2 & 26.2 & 20.0 & \textbf{25.4} & 31.2 & 22.8 & 28.8 & 19.5 & \textbf{25.6} & \textbf{29.0} \\
            \textbf{MiniCPM-Llama3-V-2.5} & 31.2 & 30.8 & 28.6 & 9.2 & \textbf{24.9} & 44.6 & 44.8 & 53.5 & 15.0 & \textbf{39.5} & \textbf{22.7} \\
            \textbf{MiniCPM-V-2.6} & 53.8 & 43.3 & 33.8 & 21.7 & \textbf{38.2} & 79.6 & 62.0 & 69.3 & 25.2 & \textbf{59.0} & \textbf{48.5} \\
            \textbf{Ovis2.5-2B} & 26.2 & 42.9 & 34.3 & 21.7 & \textbf{31.3} & 35.0 & 20.8 & 40.0 & 31.0 & \textbf{31.7} & \textbf{32.7} \\
            \textbf{Ovis2.5-9B} & 39.6 & 27.9 & 33.8 & 26.7 & \textbf{32.0} & 51.2 & 24.4 & 44.2 & 37.6 & \textbf{39.4} & \textbf{37.3} \\
            \textbf{Phi-4-multimodal} & 34.6 & 26.7 & 21.0 & 35.0 & \textbf{29.3} & 42.5 & 38.4 & 39.5 & 22.7 & \textbf{35.8} & \textbf{31.0} \\
            \bottomrule[1.2pt]
        \end{tabular}
    }
    \vspace{-0.3cm}
    \caption{\textbf{MLLM performance on MIOH benchmark.} HN: Hard Negatives, HP: Hard Positives, NI: Number of Images.}
    \label{tab:performance_all}

\end{table*}
\renewcommand{\arraystretch}{1.0}

\subsection{Adversarial Pressures}
\label{sec:adversarial_pressure}

To systematically probe model vulnerabilities, we introduce three adversarial factors that create challenging but realistic evaluation conditions. These factors target distinct failure mechanisms: \textbf{visual context scale} tests integration capacity
as the number of images increases,
\textbf{perceptual difficulty} challenges feature
extraction from small or occluded objects, and \textbf{contextual bias} probes
susceptibility to misleading co-occurrence priors.

\vspace{0.1cm} \noindent
\textbf{Visual Context Scale (Number of Images).} Inspired by findings that MLLMs struggle to identify information across large image sets (the ``Visual Haystack'' problem \citep{wu2024visual}), we systematically vary the \textbf{Number of input Images (NI)} among
$\{2, 4, 8, 10\}$ for the same underlying question.
This tests the model's \emph{integration capacity}—as the visual context expands with more images, models must maintain accurate object representations across an increasing number of visual scenes.

\vspace{0.1cm} \noindent
\textbf{Perceptual Difficulty (Hard Positive).} Small or partially occluded objects are inherently harder to detect \citep{zhang2024exploring, liu2025beyond, wei2025coco}.
We curate \textbf{Hard Positive (HP)} examples using two complementary approaches:
(a) \emph{Rule-based filtering} selects images where target objects have small bounding boxes or high occlusion ratios;
(b) \emph{CLIP-based semantic filtering} identifies cases where CLIP similarity between the image and text prompt ``A photo of [object]'' is abnormally low, indicating perceptual ambiguity.
These examples test whether models can extract accurate features under perceptually challenging conditions—a necessary prerequisite for any downstream reasoning.

\vspace{0.1cm} \noindent
\textbf{Contextual Bias (Hard Negative).} Strong contextual priors can mislead models into hallucinating objects that are contextually plausible but visually absent \citep{datta2025evaluating, li2023evaluating, li2024naturalbench}.
For example, a kitchen scene may activate strong priors for ``frying pan,'' leading to false positives.
We construct \textbf{Hard Negative (HN)} examples by:
(a) estimating co-occurrence probabilities $P(\text{object}_A | \text{object}_B)$ from COCO training data and selecting images containing high-probability co-occurring objects but missing the target;
(b) applying CLIP-based semantic confusion to find images with high visual-text similarity despite object absence.
These examples test whether models rely on contextual shortcuts rather than grounded visual evidence.

\subsection{Implementation details}
\label{sec:implementation}

\textbf{Dataset Selection.} To ensure annotation quality, we use three complementary
datasets: \textbf{COCO-ReM}~\citep{singh2024benchmarking} (existence, counting)
addresses COCO's incomplete masks and missing instances via systematic
re-annotation; \textbf{PACO}~\citep{ramanathan2023paco} (attributes) provides
standardized attribute labels across object categories; \textbf{SVG}~\citep{park2025synthetic}
(spatial relationships) offers complete scene-level annotations, unlike Visual
Genome~\citep{krishna2017visual} which averages only 1.5 relationships per subject.

\noindent\textbf{Benchmark Statistics.} Three independent annotators validated all questions. The final benchmark comprises 3,484 questions across 11,732 images, balanced across tasks, reasoning patterns, and difficulty levels. Full construction details, filtering criteria, and statistics are in~\cref{appendix:dataset_selection_details}.

\begin{figure*}[t]
    \vspace{-0.3cm}
    \centering
    \includegraphics[width=0.8\linewidth]{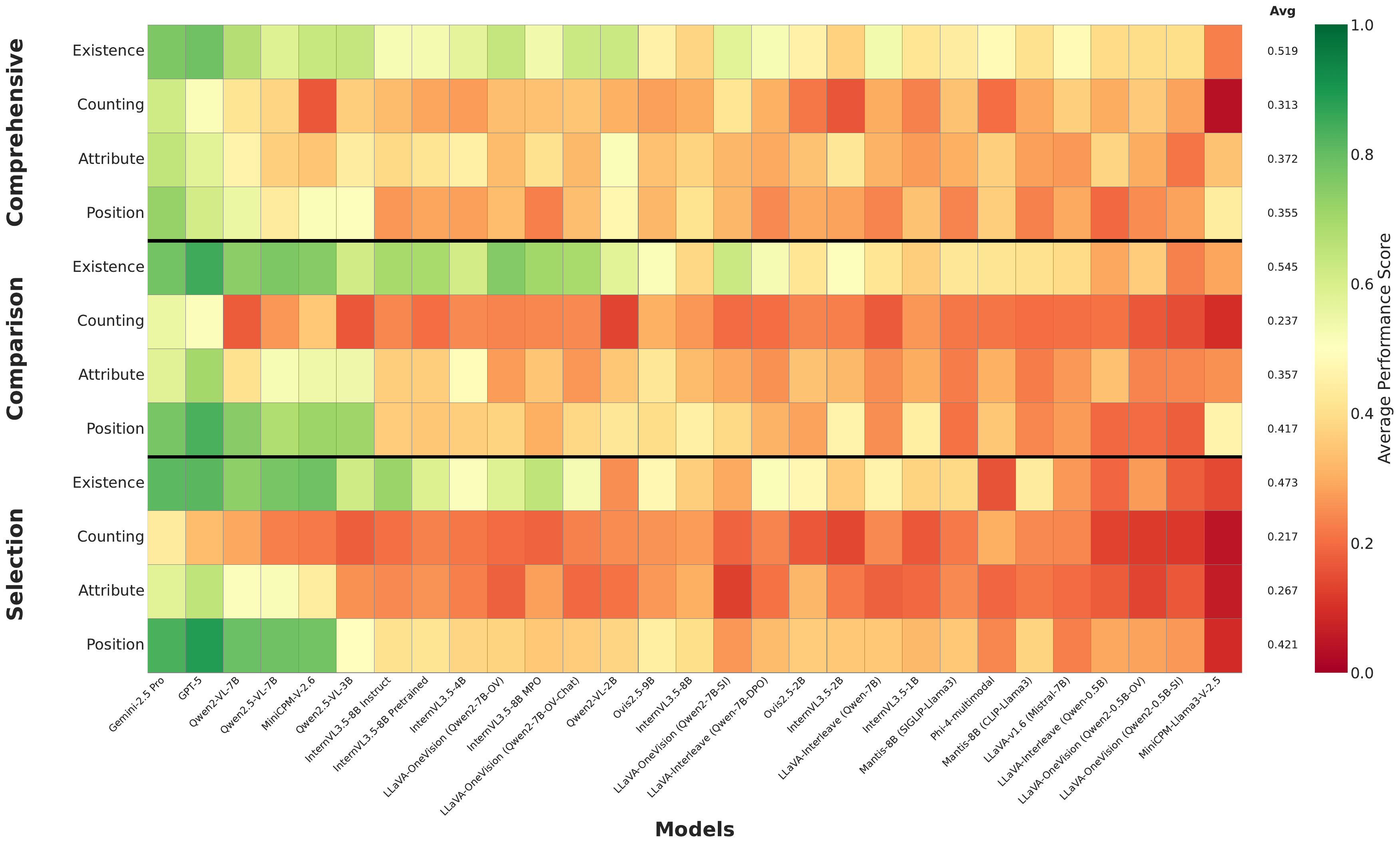}
    \vspace{-0.3cm}
    \caption{Performance heatmap across multi-image reasoning patterns (Comprehensive, Comparison, Selection) and object-centric tasks (Existence, Counting, Attribute, Position).}
    \label{fig:heatmap}
    \vspace{-0.5cm}
\end{figure*}

\section{Benchmark Results and Discussion}
\label{sec:experiments}
\vspace{-0.2cm}

We conduct a comprehensive comparative study using our MIOH over the state-of-the-art MLLMs, including GPT-5 \citep{singh2025openai} and Gemini-2.5-Pro \citep{comanici2025gemini}.
Among open-source models, we choose
the LLaVA \citep{li2024llava} series,
the Qwen \citep{bai2025qwen2} series,
InternVL \citep{wang2025internvl3},
Phi-4-multimodal \citep{abouelenin2025phi}, MiniCPM-V \citep{yao2024minicpm}, Ovis-2.5 \citep{lu2025ovis2}, and Mantis-8B \citep{jiang2024mantis}.
For reproducibility, the decoding temperature is set to 0 for all experiments.
All experiments were conducted on four NVIDIA A6000 GPUs.

\subsection{Overall Performance}
\label{sec:overall_results}

We first present overall performance results demonstrating the extent of object hallucination challenges across different model categories and tasks.
\cref{tab:performance_all} reports the comprehensive evaluation results over 29 models, revealing that multi-image object hallucination remains as significant challenge, with an overall average accuracy of only \textbf{36.1\%}.
Not surprisingly, a clear performance gap is measured between proprietary and open-source models.
The leading models, Gemini-2.5 Pro (64.4\%) and GPT-5 (63.1\%), set the state-of-the-art, but are still far from perfect.
Top-performing open-source models, such as Qwen2-VL-7B (49.1\%) and MiniCPM-V-2.6 (48.5\%), demonstrate strong capabilities but lag behind the frontier models.

Beyond the overall gap, performance is shaped by three key factors—reasoning pattern, task type, and adversarial pressure—each revealing distinct failure modes, which we analyze in the following subsections.

\subsection{Reasoning Patterns}

Our analysis reveals that MLLM performance is governed not only by the task but also, fundamentally, by the required reasoning pattern. As shown in \cref{fig:heatmap}, performance varies significantly across Comprehensive, Comparison, and Selection patterns, demonstrating that existing object hallucination benchmarks, constrained to a few question styles, cannot properly assess object hallucination across diverse multi-image settings.

Overall, Comprehensive emerges as the most manageable reasoning pattern, with basic Existence questions yielding the highest performance. Comparing this densely high-performing row against the rest of the heatmap reveals that models are heavily biased towards simple, presence-verifying queries. Evaluating models solely within this narrow scope inevitably overestimates their true robustness.

On average, Selection(34.2\%) emerges as the most challenging reasoning pattern, underperforming both Comprehensive and Comparison. This difficulty becomes especially pronounced when paired with compositional tasks such as Attribute(26.7\%), indicating that models often recognize that an object-attribute pair exists within the image set but fail to identify which specific image contains it—a distinct form of grounding failure in multi-image settings.

\subsection{Task-Specific Vulnerability}

We conduct detailed analyses to identify specific failure patterns and characterize how different adversarial pressures affect model behavior in multi-image scenarios.
As reported in \cref{tab:performance_all}, the overall task-level accuracy reveals a clear hierarchy of difficulty.
Existence (49.4\%) emerges as the most manageable task for most models, suggesting a basic level of object recognition.
In stark contrast, counting (25.4\%) is a critical failure point across the board, highlighting a fundamental weakness in quantitative reasoning under multi-image contexts.
Attribute (32.4\%) and position (37.3\%) tasks reveal significant difficulty as well, underscoring the challenges of compositional understanding.

\vspace{0.1cm} \noindent \textbf{Existence.}
While Existence is the highest-scoring task, model performance is brittle.
The accuracy on easy samples (62.4\%) consistently drops on perceptually challenging objects (HP, 51.5\%) or on contextually misleading scenes (HN, 51.9\%).
This indicates that models' understanding of object presence is heavily reliant on clear, unambiguous visual cues and can be easily disrupted, leading to false negative hallucinations.

\vspace{0.1cm} \noindent
\textbf{Counting.}
The average accuracy of 25.4\% confirms that counting is a core deficiency.
Our sub-task analysis reveals this is not limited to one failure mode; models struggle with both aggregating counts across images
and locating an image with a specific number of objects.
This points to a fundamental inability in quantitative grounding,
especially on multiple scenes.

\vspace{0.1cm} \noindent
\textbf{Attribute and Position.}
These tasks require binding objects to their properties or spatial relations, a challenge of compositional reasoning.
The low accuracy in the attribute task (32.4\%) suggests that MLLMs may recognize an object and an attribute separately but fail to confirm their visual co-occurrence (\textit{e.g.}, seeing a `car' and `red' but hallucinating a `red car').
Similarly, the position task scores slightly higher (37.3\%), but degrades under Hard Negative pressure, indicating that models often tend to overlook complex spatial relationships beyond mere existence.

\subsection{Impact of Adversarial Pressures}

As reported in \cref{tab:performance_all}, introducing adversarial constraints consistently degrades MLLM performance across all tasks relative to their easy baselines.

\vspace{0.1cm} \noindent \textbf{Perceptual and Contextual Pressures (HP \& HN).}
Performance on Existence, Attribute, and Position consistently drops under both HP and HN conditions.
For instance, Attribute accuracy falls from 37.5\% to 31.3\% (HP), and Position from 46.9\% to 35.2\% (HN), confirming model vulnerabilities to misleading visual and contextual cues. In contrast, Counting exhibits only a marginal decline.
However, this reflects a \textit{floor effect} rather than genuine robustness; with baseline counting accuracy already near random chance (25.4\%), added complexities leave little room for further degradation.

\vspace{0.1cm} \noindent \textbf{Cognitive Load from Multiple Images (NI).}
While HP and HN induce moderate decays, increasing the number of input images to 8 (NI) triggers a far more severe collapse. Under this heightened cognitive load, even the fundamental Existence task plummets to 30.0\%—a relative drop of exceeding 50\% from the easy baseline. Although top-tier models like GPT-5 and Gemini 2.5 Pro manage this expanded context better than average, their substantial performance drops confirm that scaling multi-image reasoning remains a critical open challenge.

\subsection{Accuracy-Robustness Trade-off}

\begin{figure}[t]
    \centering
    \includegraphics[width=\columnwidth]{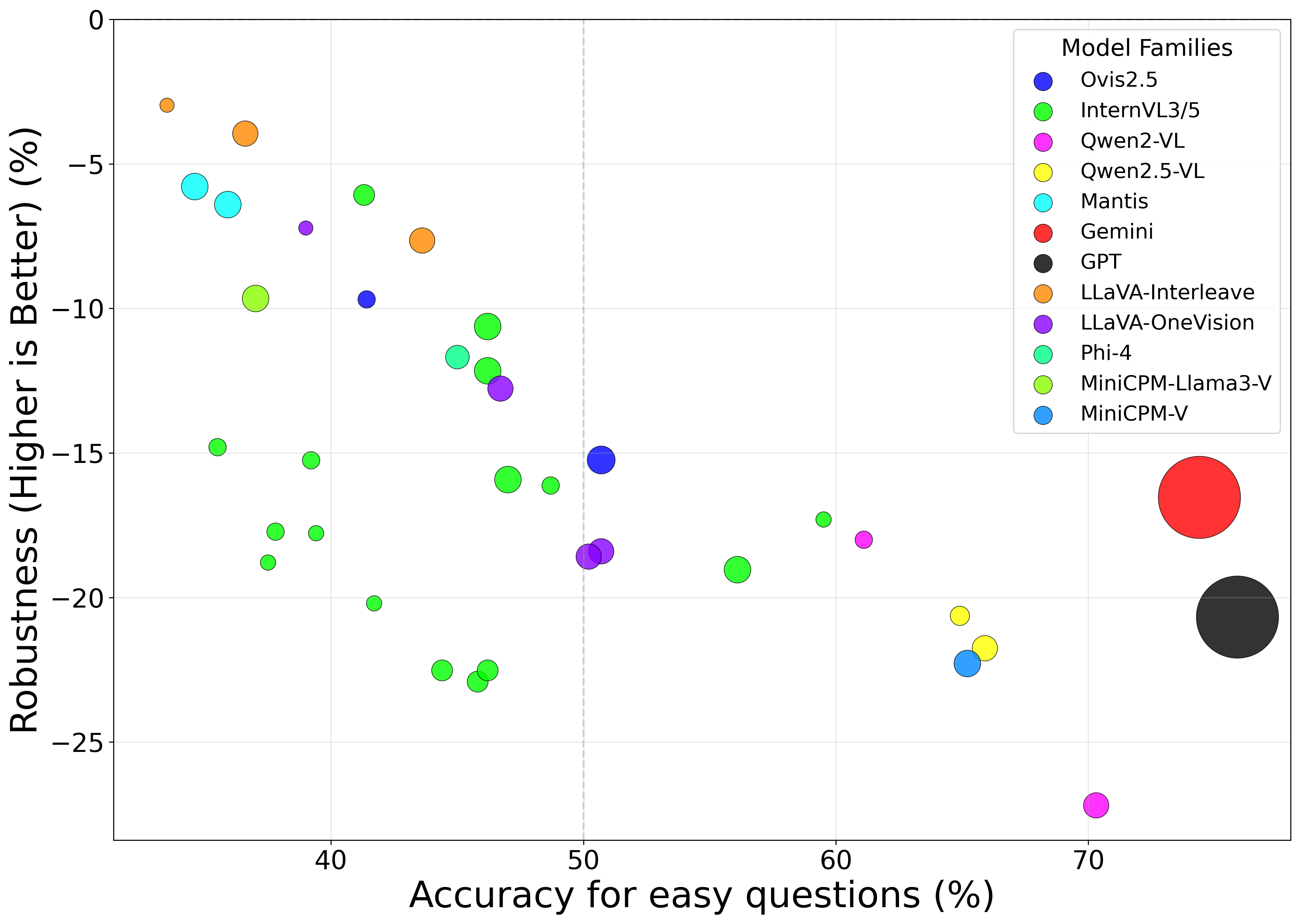}
    \caption{\textbf{Accuracy on easy questions and robustness under adversarial conditions in VLMs.}
    Circle size indicates model size, where Gemini and GPT size is based on estimation.}
    \label{fig:easy_vs_robustness}
\end{figure}

As shown in \cref{fig:easy_vs_robustness},  our analysis uncovers a fundamental trade-off between achieving high accuracy on straightforward tasks and maintaining robust performance under adversarial pressures.
In other words, a strong baseline performance does not guarantee robustness against object hallucination.
Open-source models with higher baseline accuracy often exhibit greater vulnerability to adversarial conditions.
The correlation analysis reveals a moderate positive relationship between model size and performance on easy questions, but virtually no correlation between size and robustness, suggesting that simply scaling model parameters does not inherently improve resilience to object hallucination.
The top-performing open-source models on easy questions—Qwen2-VL-7B and Qwen2.5-VL-7B—experience substantial robustness drops of -27.2\% and -21.8\%, respectively, indicating that high capability models might be more susceptible to the adversarial pressures we designed.
This finding suggests that the ability to handle straightforward multi-image tasks does not guarantee robustness against object hallucination.

\begin{figure}[t]
    \centering
    \includegraphics[width=\columnwidth]{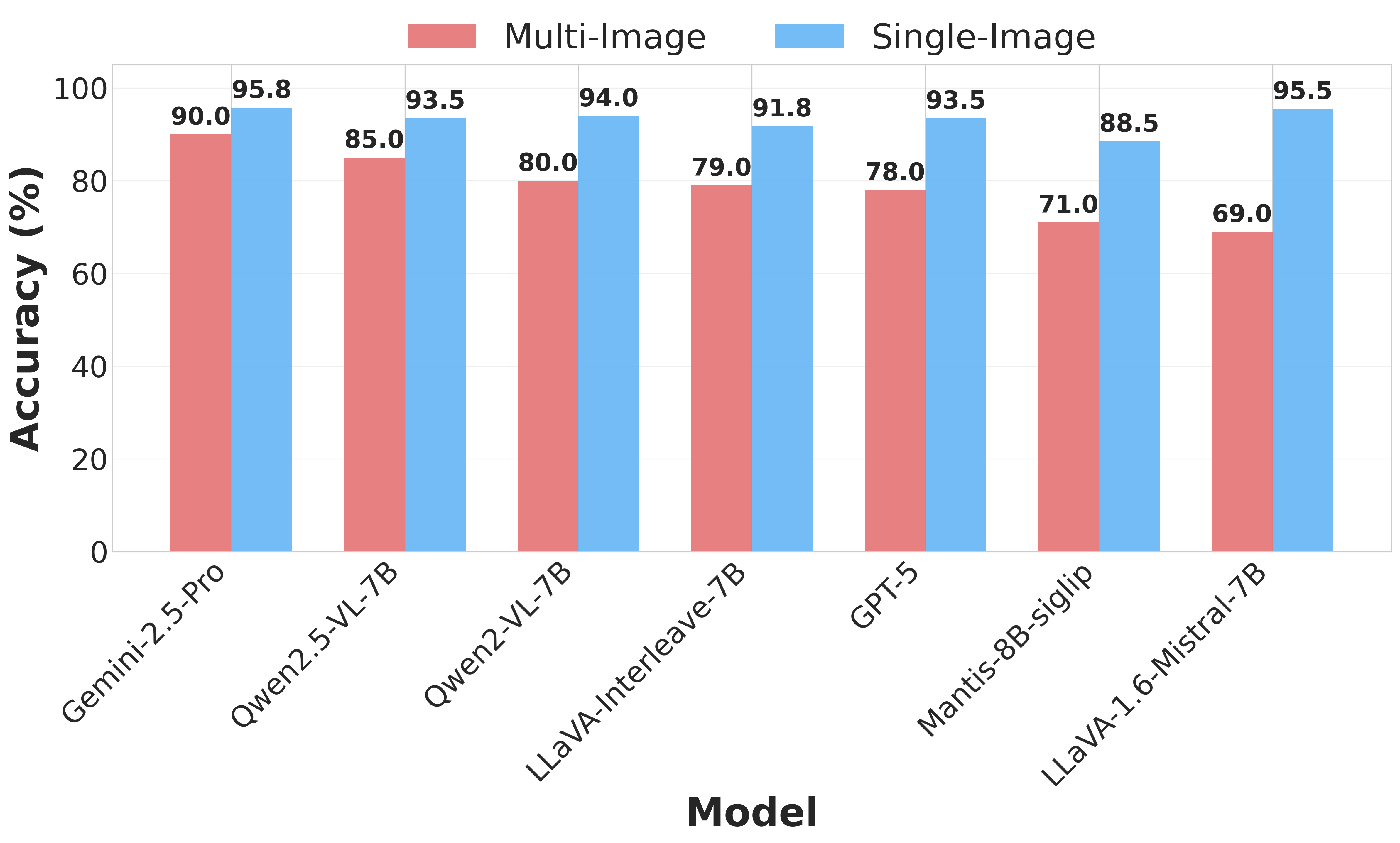}
    \caption{\textbf{Multi-image processing consistently degrades object hallucination across all models.}
    Comparison of accuracy between multi-image comprehensive (red bars) and single-image decomposed (blue bars) evaluation on Existence task.}
    \label{fig:aggregaat}
\end{figure}

\subsection{Multi-Image Context as a Hallucination Amplifier: An Ablation Study}

To isolate the impact of multi-image processing on object hallucination, we conduct a controlled ablation study focusing on the Existence task.
Specifically, we compare two evaluation approaches for identical visual content: \textbf{comprehensive} questions that require synthesizing information across all images simultaneously (``Is there an \textsc{object} in any of these \textsc{image}s?'') \textit{vs}. \textbf{decomposed} questions that mirror the traditional single-image setting by asking each image separately (``Is there a \textsc{object} in \textsc{image}?'') and combining the answers to determine overall presence.
This design isolates whether multi-image contexts introduce systematic errors beyond simple accumulation of individual image processing mistakes.

The results in \cref{fig:aggregaat} reveal that single-image processing substantially outperforms simultaneous multi-image analysis, consistently across all models and scales.
This indicates that multi-image contexts significantly amplify object hallucination beyond what error accumulation alone would predict, pointing to cross-image integration as a primary source of failure. The consistency of this penalty suggests that current MLLM training paradigms fail to adequately address object hallucination in multi-image reasoning.

\section{Conclusion}
\label{sec:conclusion}

We introduce MIOH, a benchmark that systematically evaluates object hallucination in multi-image contexts across four object-centric tasks (Existence, Counting, Attribute, Position) with three reasoning patterns (Comprehensive, Comparative, Selective) under three adversarial pressures (Visual Context Scale, Perceptual Difficulty, Contextual Bias).
Through evaluation of 29 models, we demonstrate that current MLLMs—including GPT-5 and Gemini-2.5-Pro—exhibit fundamental limitations in multi-image integration.
Our fine-grained diagnostic framework reveals that hallucination stems not merely from perceptual failures but from integration-stage breakdowns when maintaining object representations across multiple images.
MIOH provides a foundation for targeted improvements in multi-image understanding and serves as a critical evaluation tool for developing more reliable multimodal AI systems.

\section*{Acknowledgments}
This work was also supported by the SOFT Foundry Institute at SNU, Samsung Electronics, Youlchon Foundation, National Research Foundation of Korea (NRF) grants (RS-2021-NR05515, RS-2024-00336576, RS-2023-0022663, RS-2025-25399604, RS-2024-00333484), and the Institute for Information \& Communication Technology Planning \& Evaluation (IITP) grants (RS-2022-II220264, RS-2024-00353131) funded by the Korean government.

{
    \small
    \bibliographystyle{ieeenat_fullname}
    \bibliography{main}
}

\clearpage
\clearpage
\setcounter{page}{1}
\maketitlesupplementary

\appendix

\pagenumbering{roman}
\renewcommand\thetable{\Roman{table}}
\renewcommand\thefigure{\Roman{figure}}
\setcounter{page}{1}
\setcounter{section}{0}
\setcounter{table}{0}
\setcounter{figure}{0}

\section{Detailed Benchmark Construction} \label{appendix:dataset_selection_details}
In this section, we provide comprehensive details on our benchmark construction process, including: the curation and selection criteria for our datasets (\cref{appendix:curation}), the metadata construction methodology (\cref{appendix:metadata_construction_details}), the automated question generation framework (\cref{appendix:question_generation}), and the adversarial benchmark design strategy (\cref{appendix:adversarial}).

\subsection{Dataset Curation}
\label{appendix:curation}

\subsubsection{COCO-ReM}

\textbf{Annotation Quality Issues in Original COCO.} The original COCO dataset annotations contain significant gaps that make them unsuitable for reliable object hallucination evaluation. These issues include incomplete object masks, missing instances, and inaccurate bounding boxes that would introduce systematic errors in our rule-based question generation framework.

\noindent\textbf{COCO-ReM Improvements.} COCO-ReM (Refined Masks)~\citep{singh2024benchmarking}  addresses these limitations through a comprehensive re-annotation process: (1) \textit{Mask boundary refinement} using the Segment Anything Model (SAM) to improve precision, (2) \textit{Missing instance detection} using advanced detection models to identify previously unlabeled objects, (3) \textit{Label correction} through systematic review and human validation, and (4) \textit{Enhanced object masks and bounding boxes} providing more complete scene coverage.

\noindent\textbf{Impact on Benchmark Quality.} As demonstrated in RePOPE~\citep{neuhaus2025repopeimpactannotationerrors}, high-reliability annotations significantly impact ground truth accuracy, making this a crucial consideration for benchmark design. The enhanced annotation quality in COCO-ReM ensures our existence and counting questions have reliable ground truth labels, substantially reducing false negatives that could arise from missed objects in original COCO annotations.

\noindent\textbf{Object Count Limitations.} During validation, we observed that even COCO-ReM's accuracy degrades when object counts exceed certain thresholds. Specifically, images containing more than 10 objects showed decreased annotation reliability. To maintain benchmark integrity, we implemented a conservative approach by limiting counting questions to images with 5 or fewer objects, ensuring high reliability through validated annotations while preserving sufficient complexity for meaningful MLLM evaluation.

\subsubsection{PACO}

\textbf{Limitations of Existing Attribute Datasets.} While various datasets address object attributes, they suffer from systematic limitations: (1) Original COCO annotations lack standardized attribute labeling across object categories, (2) COCO Attributes~\citep{patterson2016coco} provides standardized annotation but suffers from limited diversity in both object categories and attribute types, and (3) Insufficient coverage for comprehensive benchmark construction requiring comparison across diverse objects and attributes.

\noindent\textbf{PACO's Comprehensive Approach.} PACO (Parts and Attributes of Common Objects)~\citep{ramanathan2023paco} provides a superior solution through: (1) Broader category coverage spanning a more diverse range of object types, (2) Systematic attribute annotation ensuring consistency across identical objects, (3) Detailed annotation process that identifies constituent object parts and labels their diverse attributes, and (4) Large-scale structured dataset resulting in comprehensive fine-grained object understanding capabilities.

\noindent\textbf{Advantages for Question Generation.} PACO's structured approach offers several key benefits: systematic attribute labeling with sufficient scale and diversity to support robust question generation, extensive object-attribute combinations enabling comprehensive evaluation across diverse visual scenarios, standardized annotation framework ensuring consistent evaluation criteria across different object categories, and high-quality ground truth reducing ambiguity in attribute-based question validation.

\subsubsection{SVG}

\textbf{Limitations of Existing Spatial Relation Datasets.} Existing datasets for spatial relationship evaluation suffer from critical annotation gaps: Visual Genome~\citep{krishna2017visual} and GQA~\citep{hudson2019gqa} provide relation data but have incomplete spatial relationship coverage, missing relationships in ground truth annotations that exist visually but are not labeled, and annotation inconsistencies that reduce reliability for systematic evaluation.

\noindent\textbf{SVG's Multifaceted Approach.} SVG (Synthetic Visual Genome)~\citep{park2025synthetic} addresses these limitations through comprehensive methodology: object detection integration for accurate entity identification, scene graph enhancement to capture missing relationships, region descriptions providing contextual relationship validation, depth information enabling more accurate spatial reasoning, region masks for precise relationship localization, VQA-based verification for non-spatial relationships to ensure annotation quality, and systematic filtering to reduce incorrect relationship annotations.

\noindent\textbf{Key Advantages for Spatial Evaluation.} SVG provides several critical improvements: (1) Richer spatial relation coverage per subject compared to existing datasets, enabling more comprehensive spatial reasoning evaluation, (2) Comprehensive filtering that systematically reduces incorrect relationships, improving ground truth reliability, (3) Region mask-based verification enabling more reliable relationship identification through visual evidence, and (4) High relation density minimizing the critical impact of missing positional relationships on question accuracy. These enhancements make SVG particularly well-suited for generating position-based questions that can reliably assess MLLM spatial reasoning capabilities in multi-image contexts, where accurate relationship identification becomes even more challenging due to increased visual complexity.

\subsection{Metadata Construction} \label{appendix:metadata_construction_details}
Having established our data sources, we now detail the metadata construction process that enables efficient question generation.

\noindent\textbf{Hierarchical Organization Structure.} Our metadata follows a systematic three-level organization: (1) \textit{Task-specific property categorization} where objects are categorized by relevant attributes, relations, or counts, (2) \textit{Difficulty level classification} with Easy/Hard Negative/Hard Positive assignments based on visual and semantic complexity, and (3) \textit{Image identifier mapping} where specific image IDs are linked to categorized objects for efficient retrieval.

\noindent\textbf{Rule-Based Filtering Criteria.} We implement several filtering mechanisms: minimum bounding box size requirements to ensure object visibility, occlusion level thresholds based on mask overlap calculations, and image resolution considerations for consistent object detectability across different image qualities. For difficulty classification, we define easy positives/negatives as clear, unambiguous cases with high visibility and minimal contextual confusion, hard positives as present objects with small size, high occlusion, or minimal contextual cues, and hard negatives as absent objects in contexts with high co-occurrence bias or semantic similarity.

\noindent\textbf{CLIP-Based Semantic Similarity Implementation.} Our similarity score calculation involves text prompt generation using standardized formats (``A photo of [object]'',``[attribute] [object]''), image encoding through CLIP visual encoder, cosine similarity computation between text and image embeddings, and threshold determination through empirical validation on representative samples. This metadata system enables rapid question synthesis while maintaining quality through automated filtering based on rule-based criteria, semantic validation using CLIP similarity scores, systematic difficulty categorization across different visual reasoning scenarios, and efficient question generation through pre-computed metadata lookup.

\noindent\subsection{Question Generation}
\label{appendix:question_generation}

Using the prepared positive and negative samples per task, we generate multi-image questions that instantiate the three types of questions (comprehensive, comparative, and selective) across multi-image contexts for each of the four core tasks.
As all ingredients are ready, this step can be easily automated by rule-based templates; \emph{e.g.}, for Counting, we construct a question by randomly selecting $N$ \textsc{image}s containing a particular \textsc{object}, and the sum of \textsc{count} per each example is the right answer.
Other tasks follow a similar procedure to generate the question and correct answer.
All questions are constructed in multiple-choice (MCQ) format by incorporating incorrect answers.
We also include the ``None of the above'' options to more precisely assess the model's understanding.

\noindent\subsection{Adversarial Example Construction}
\label{appendix:adversarial}

Building upon the adversarial pressures described in ~\cref{sec:adversarial_pressure},
we detail the implementation procedures for constructing challenging evaluation examples.

\subsubsection{Hard Positive Example}
We employ two complementary filtering approaches to identify perceptually difficult positive examples: \textbf{Rule-based filtering} selects (\textsc{image}, \textsc{object})
pairs where the target object exhibits challenging visual characteristics.
We filter based on bounding box dimensions relative to image size,
segmentation mask area indicating occlusion levels, and spatial positioning
within the image frame. This approach captures objects that are inherently
difficult to perceive due to size or visibility constraints.
\textbf{CLIP-based semantic filtering} identifies cases where visual-semantic
alignment is weak. We compute similarity scores between image embeddings and
text prompts formatted as ``A photo of \textsc{object}'' using CLIP.
Examples with unexpectedly low similarity scores despite object presence indicate
perceptual ambiguity—such as unusual viewpoints, partial visibility, or
atypical visual contexts that challenge standard recognition patterns.

\subsubsection{Hard Negative Example}
We construct misleading negative examples through two strategies:
\noindent\textbf{Co-occurrence-based selection} leverages statistical patterns from
COCO training data. We compute pairwise co-occurrence probabilities
$P(\text{object}_A | \text{object}_B)$ across all object categories.
For a target object, we select images containing frequently co-occurring
objects while the target itself is absent, creating scenarios where strong
contextual priors may mislead models into false positive predictions.
\noindent\textbf{CLIP-based semantic confusion} identifies images that exhibit
high visual-semantic similarity with target object prompts despite the object's
absence. We compute CLIP similarity between images and ``A photo of \textsc{object}''
prompts for absent objects, selecting cases with high similarity scores.
These examples represent strong false association triggers where visual context
strongly suggests object presence without actual visual evidence.

\begin{figure*}
    \vspace{-0.5cm}
    \centering
    \includegraphics[width=\linewidth]{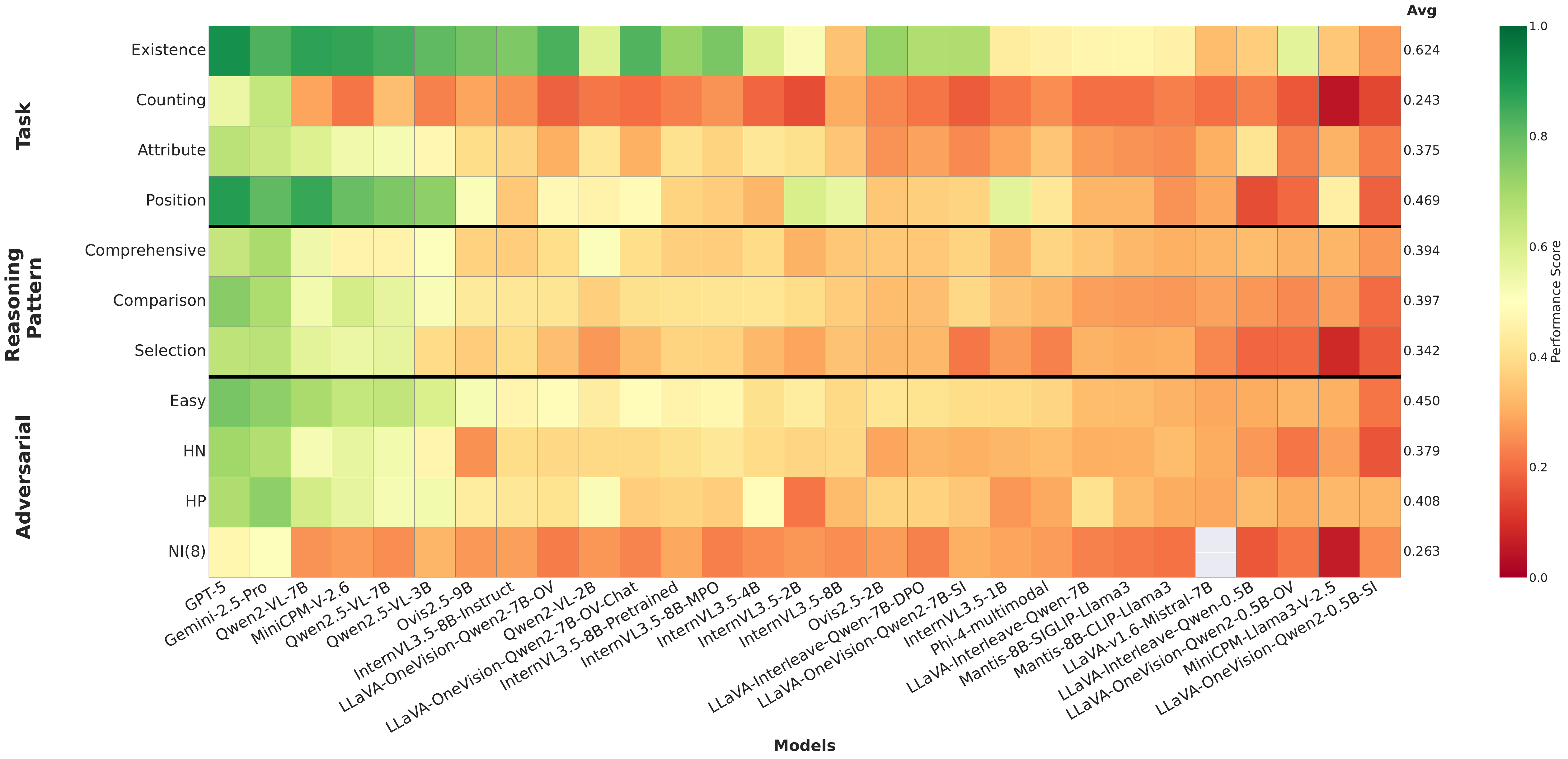}
    \caption{\textbf{Aggregated Performance Heatmap across Task, Reasoning Pattern, and Adversarial Pressure Dimensions.}}
    \label{fig:advanced_heatmap}
    \vspace{-0.3cm}
\end{figure*}

\subsubsection{Difficulty Level Integration.}
During question generation, we
systematically control difficulty by varying the proportion of challenging
examples. For a question requiring $N$ images, we can include anywhere from
0 to $N$ hard positive or hard negative examples, with difficulty increasing
as the proportion of challenging examples approaches $N$.

\subsubsection{Quality Assurance and Validation}
To ensure benchmark reliability, we conducted comprehensive manual validation
and systematic sampling to construct a balanced evaluation set.
Our automated generation pipeline initially produced over 26,000 questions
across all task types and reasoning patterns.

Each question underwent independent review by three annotators with expertise in computer vision
and visual reasoning tasks. Annotators assessed: (1) ground truth correctness
based on visual evidence, (2) question clarity and lack of ambiguity,
(3) validity and distinctiveness of multiple choice options, and
(4) consistency with source dataset annotations.
Disagreements were resolved through majority voting, with persistent ambiguities
leading to question removal.

Despite carefully selecting well-annotated datasets for construction, the validation process revealed that certain task types were still particularly prone to systemic issues. Position questions frequently exhibited ambiguous or underspecified spatial relationships, often stemming from incomplete or inconsistent relationship annotations in the source datasets. Similarly, Counting questions continued to suffer from missed instances or miscounted objects, indicating that even high-quality datasets contain non-trivial annotation noise for fine-grained quantitative reasoning. After validation and filtering, approximately 20,000 questions remained as the verified question pool.

To ensure fair and comprehensive evaluation across all dimensions, we applied
stratified sampling to construct the final benchmark with balanced distribution
across three key axes: Task Types, Reasoning Patterns, and Adversarial Pressures.
This sampling procedure resulted in the final benchmark comprising 3,484
questions across 11,732 images, ensuring both high-quality ground truth
through rigorous validation and fair evaluation coverage across all
benchmark dimensions.

\section{Detailed Performance Analysis}
\label{appendix:analysis}

In this section, we present a comprehensive performance analysis of various multimodal large language models (MLLMs). Before delving into the complex interactions between different factors, we first examine the aggregated performance of models across three primary dimensions: \textbf{Task}, \textbf{Reasoning Pattern}, and \textbf{Adversarial Pressure.}
\cref{fig:advanced_heatmap} illustrates the performance overview.
This visualization allows for a direct comparison of how different models handle specific types of challenges independently.

\begin{figure*}
    \vspace{-0.5cm}
    \centering
    \includegraphics[width=\linewidth]{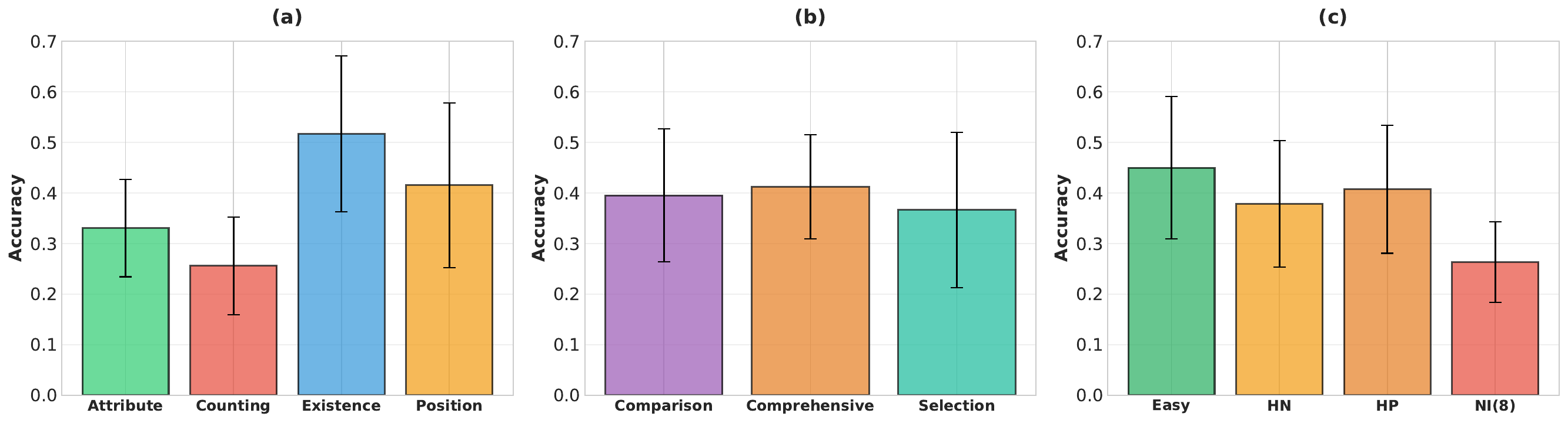}
    \caption{\textbf{Performance mean and variance.} (a) Tasks, (b) Reasoning patterns, and (c) Adversarial Pressures.}
\label{fig:marginal}
\end{figure*}

\begin{figure*}
    \centering
    \includegraphics[width=0.87\linewidth]{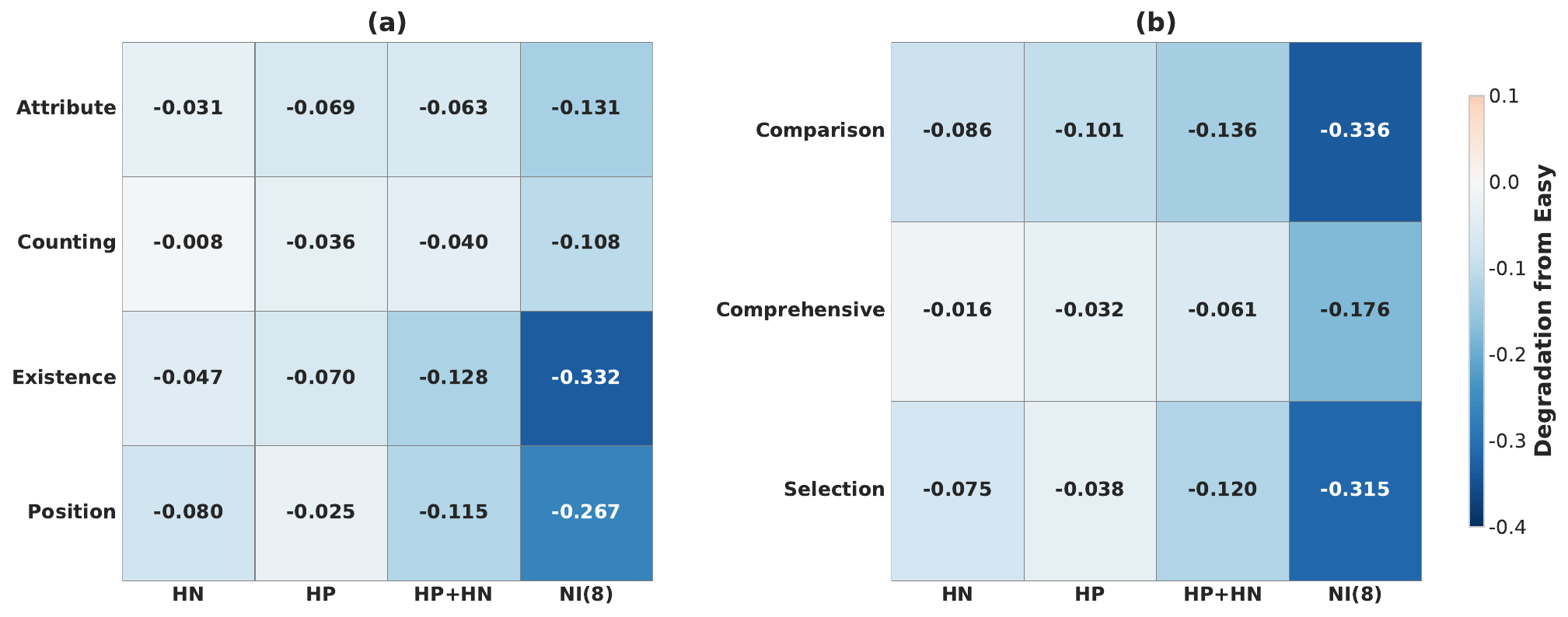}
    \caption{\textbf{Cross-dimensional degradation analysis.}
  Performance degradation from Easy baseline across (a) Task types and (b) Reasoning dimensions under adversarial pressure.}
\label{fig:combined_heatmap}
\vspace{-0.3cm}
\end{figure*}

\subsection{Variation Analysis}
\label{appendix:varinace}

\cref{fig:marginal} presents marginal performance distributions across each dimension, illustrating their capacity to distinguish model capabilities. Among task types, Position tasks demonstrate the highest variance ($\sigma = 0.163$), suggesting that models exhibit notably different levels of spatial understanding ability.
In contrast, Counting tasks show the lowest variance ($\sigma = 0.096$), indicating that this task type presents consistent difficulty across most models.
Similarly, among adversarial pressure conditions, increasing Number of Images(NI) exhibits the lowest variance ($\sigma = 0.079$), comparable to Counting tasks, suggesting that extreme context pressure leads to uniformly poor performance across models.
For reasoning patterns, Selection demonstrates the highest variance ($\sigma = 0.153$), suggesting that the ability to accurately identify a specific image among multiple candidates varies considerably across different models.

\subsection{Cross-dimensional Interaction Analysis}
  \label{appendix:interaction}
  \cref{fig:combined_heatmap} presents degradation heatmaps showing performance drops from the Easy baseline across different task-pressure and reasoning-pressure combinations, averaged over all models. Increasing Number of Images(NI) causes catastrophic degradation across all dimensions, with Comparison reasoning and Existence tasks  most severely affected.
  Combined(HP+HN) pressure shows additive difficulty, causing larger drops than either pressure alone. Comprehensive reasoning demonstrates superior robustness compared to Comparison and Selection, suggesting holistic reasoning strategies better withstand adversarial pressure.
  Counting tasks show minimal degradation not due to robustness but floor effects, as their Easy baseline is already low. These patterns reveal that adversarial robustness is highly dimension-dependent.

\subsection{Performance Gap Between Open-Source and Commercial Models}
\label{appendix:sota-comparison}
\begin{figure}
    \centering
\includegraphics[width=\linewidth]{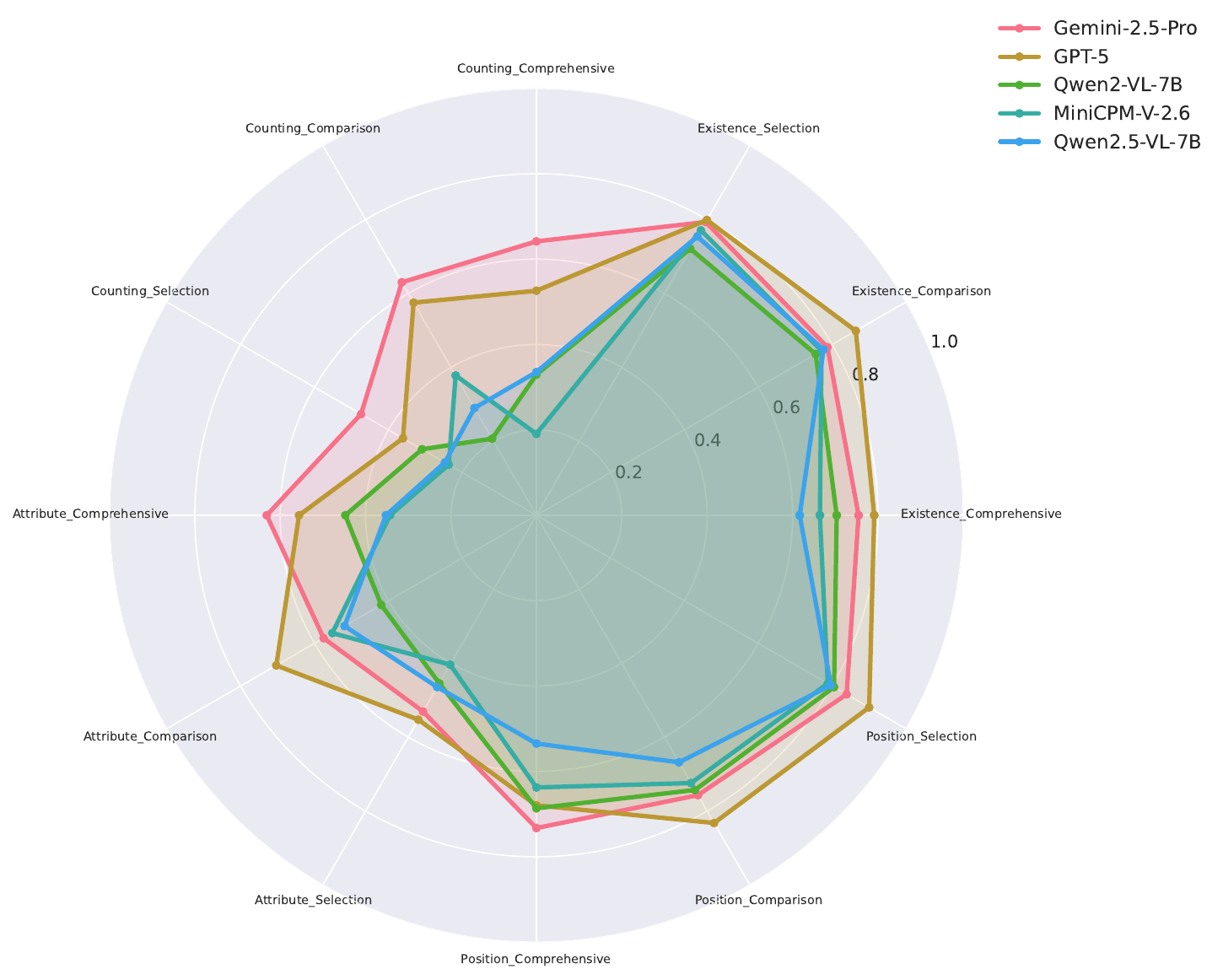}
    \caption{\textbf{Performance of commercial and top open-source models.}}
\label{fig:radar}
\end{figure}

We compare the capabilities of leading commercial models and top-performing open-source models in \cref{fig:radar}. The performance gap between these two categories is most pronounced in Counting and Comparison tasks. These capabilities represent the areas where commercial models demonstrate the largest advantages over open-source alternatives, indicating differences in multi-image reasoning capacity.

\begin{figure*}[t]
    \centering
    \includegraphics[width=\linewidth]{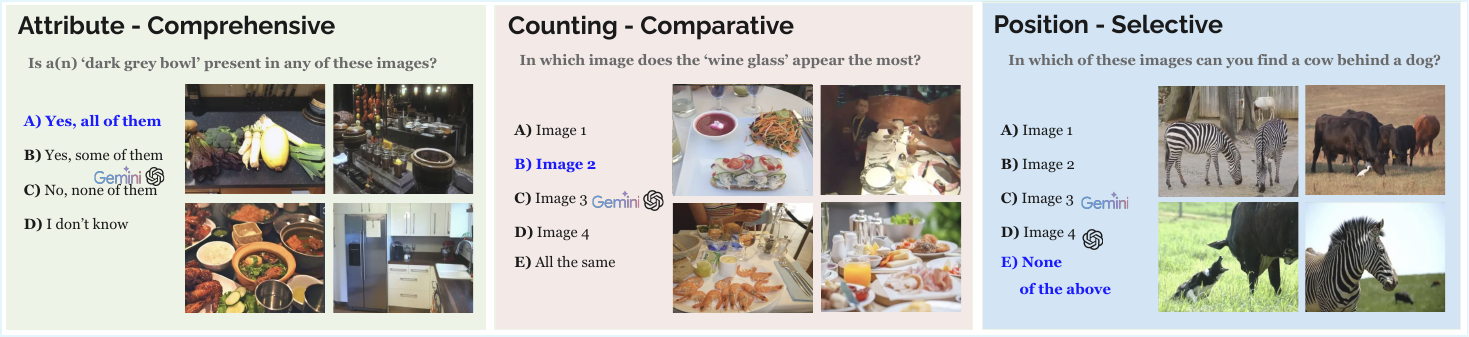}
    \caption{\textbf{Failure cases of top commercial models} (GPT-5, Gemini-2.5-pro). Blue text indicates the Ground Truth (GT). The icons indicate the incorrect choices made by each model. These examples highlight how perceptual difficulty lead to reasoning failures.}
    \label{fig:sup:appendix_failurecase}
\end{figure*}

\section{Benchmark Examples}
\label{appendix:examples}

\subsection{Qualitative Examples from MIOH benchmark}
\begin{figure*}
    \centering
    \includegraphics[width=\linewidth]{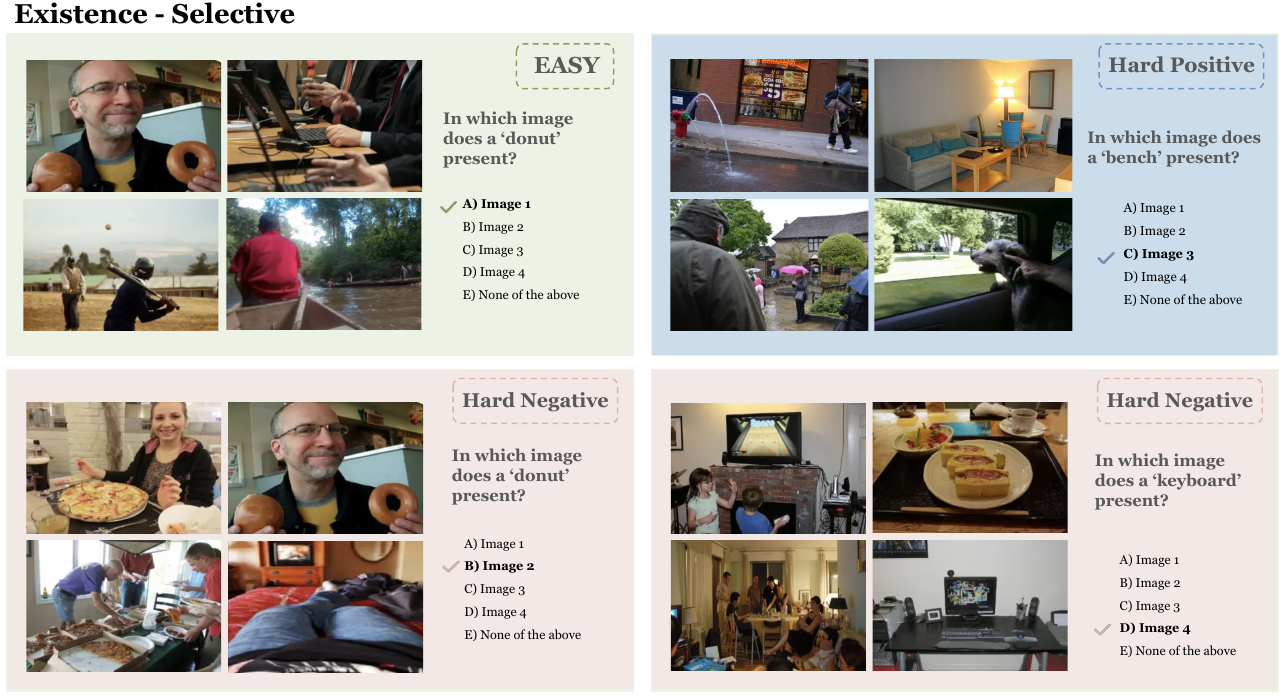}
    \includegraphics[width=\linewidth]{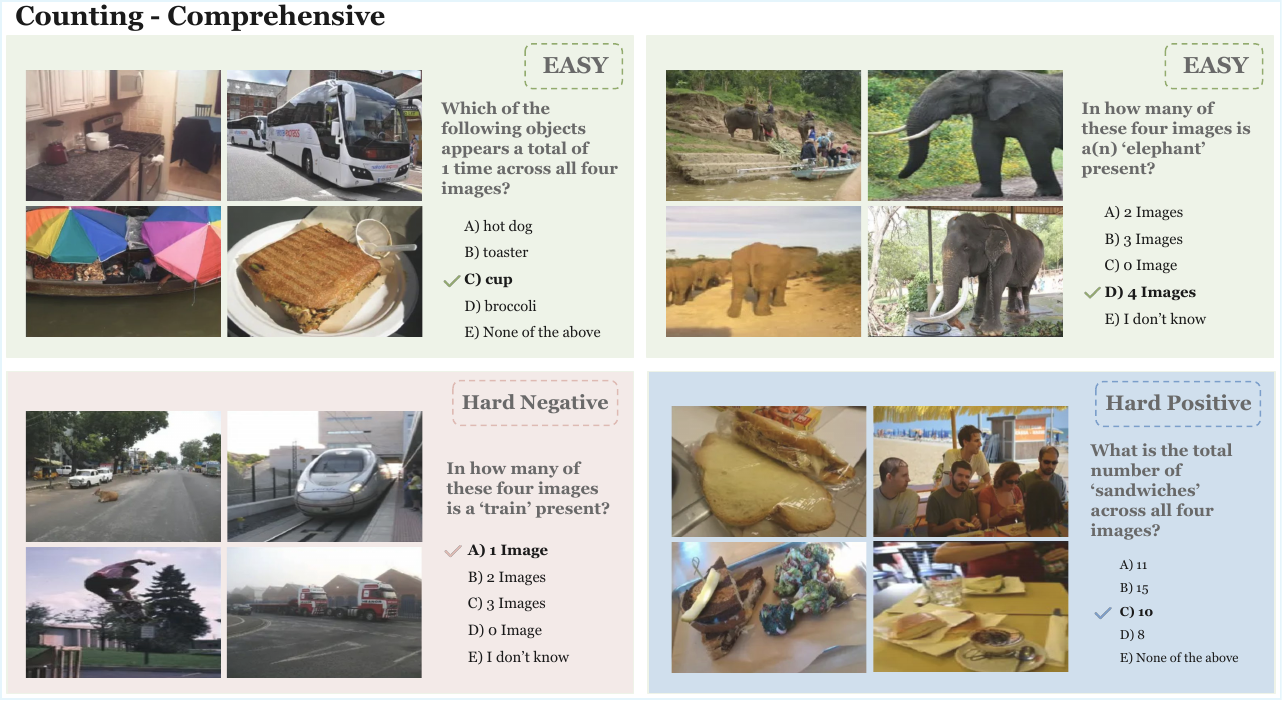}
    \caption{\textbf{Benchmark Examples 1.} Existence and Counting Task}
    \label{fig:sup:appendix_exp1}
\end{figure*}

\begin{figure*}
    \centering
    \includegraphics[width=\linewidth]{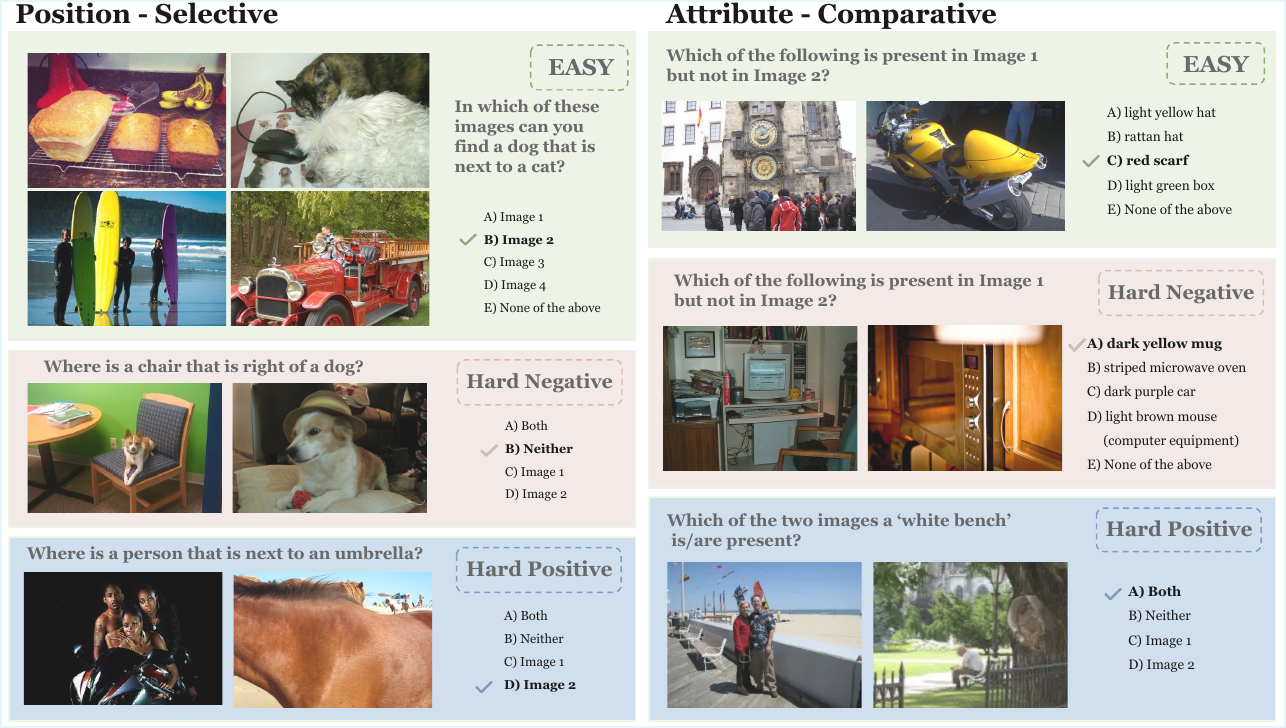}
    \caption{\textbf{Benchmark Examples 2.} Position and Attribute Task}
    \label{fig:sup:appendix_exp2}
\end{figure*}

\cref{fig:sup:appendix_exp1,fig:sup:appendix_exp2} provide qualitative examples from the MIOH benchmark, illustrating how questions are formulated across our four core tasks and various visual adversarial pressures. Each example is designed to test a specific aspect of an MLLM's object-centric capabilities and robustness against perceptual challenges.

\textbf{Existence Tasks.} (\cref{fig:sup:appendix_exp1}, top section) assess the model's ability to verify the presence of objects and pinpoint their location within the image set. The \textit{Selective} questions showcased here require the model to identify the specific image containing the target object among candidates. Examples progress from straightforward cases (Easy: clearly visible ``donut'') to perceptually challenging scenarios.
The hard positive example requires detecting a ``bench'' in a rainy scene, where the object is situated near the greenery and partially obscured by a person with umbrella, making it difficult to spot. The Hard Negative example tests robustness against contextual bias, such as identifying a ``keyboard'' in a set of images containing office-like environments or other electronics (e.g., game consoles) that visually resemble the target context but lack the specific object.

\textbf{Counting Tasks.} (\cref{fig:sup:appendix_exp1}, bottom section) evaluate quantitative reasoning capabilities through \textit{Comprehensive} questions that require aggregating counts across all provided images. The Easy examples involve basic enumeration, such as counting the occurrences of clearly visible ``elephant''s across four frames. The Hard Positive scenario challenges the model to sum the total number of ``sandwiches'' across images where objects are heavily occluded, cut into pieces, or cluttered on plates, testing the ability to handle dense visual information. The Hard Negative example asks for the count of ``trains,'' where models must distinguish the actual object from contextually similar background elements like tracks or station platforms in the non-target images.

\textbf{Position Tasks.}(\cref{fig:sup:appendix_exp2}, left section) present the most complex spatial relationship challenges. We use \textit{Selective} questions to ask the model to identify the image where a specific relation holds. Examples include Easy scenarios (``dog next to cat''), Hard Negative cases (``chair positioning relative to dog''), and Hard Positive examples (``person next to umbrella'') that test compositional scene understanding beyond simple object detection.

\textbf{Attribute Tasks.} (\cref{fig:sup:appendix_exp2}, right section) assess detailed compositional understanding by requiring models to bind visual properties with objects using \textit{Comparative} questions. The tasks range from detecting visually distinct attributes (Easy: ``red scarf'') to more subtle distinctions. The Hard Negative scenario involves identifying a ``dark yellow mug'' in a cluttered indoor environment where lighting and other objects may create confusion. The Hard Positive example tests the model's ability to consistently recognize a ``white bench'' across two different scenes despite variations in perspective and background.

Each example demonstrates the three question types designed for multifaceted evaluation: comprehensive (collective understanding across images), comparative (identifying differences between images), and selective (retrieving specific images matching descriptions). The progression of difficulty incorporates both visual factors (scale, occlusion, contextual bias) as detailed in \cref{sec:adversarial_pressure}.

\subsection{Failure Case Examples of Commercial Models}
\label{appendix:failure_analysis}

To better understand the limitations of commercial MLLMs, we analyze specific failure cases of GPT-5 and Gemini-2.5-Pro on the MIOH benchmark. Despite their strong baseline performance, these models still exhibit hallucination patterns under visual adversarial pressures. \cref{fig:sup:appendix_failurecase} illustrates representative error cases where models fail to ground visual evidence correctly within multi-image contexts.

For example, in the \textbf{Attribute-Comprehensive} task, models exhibit perceptual blindness or aggregation failures when objects are visually ambiguous or occluded; for instance, both models fail to confirm the presence of ``dark gray bowls'' in all target images, demonstrating a breakdown in comprehensive verification. The example in \textbf{Counting-Comparative} task reveals weaknesses in fine-grained classification, where models confuse ``wine glasses'' with perceptually similar objects like water goblets or tumblers, leading to incorrect frequency comparisons. Most critically, the \textbf{Position-Selective} task exposes severe object and relationship hallucination under the ``None of the above'' option. When asked for a ``cow behind a dog,'' models are forced into making a selection, resulting in hallucinated object identities (e.g., misidentifying a zebra as a cow in Image 4) or ignoring spatial constraints (e.g., selecting a dog-cow pair with incorrect positioning in Image 3), rather than correctly abstaining.

\end{document}